\documentclass{article}

\usepackage{microtype}
\usepackage{graphicx}
\usepackage{subfigure}
\usepackage{booktabs} % for professional tables
\usepackage{pifont}
\usepackage{transparent}
\usepackage{booktabs} % for professional tables
\usepackage{threeparttable}
\usepackage{longtable}
\usepackage{xspace}
\usepackage{multirow}

\usepackage{hyperref}

\usepackage[accepted]{icml2025}

\usepackage{amsmath}
\usepackage{amssymb}
\usepackage{mathtools}
\usepackage{amsthm}
\usepackage{setspace}
\usepackage{verbatim}

\usepackage[capitalize,noabbrev]{cleveref}

\theoremstyle{plain}

\theoremstyle{definition}

\theoremstyle{remark}

\newcommand{\boldres}[1]{{\textbf{\textcolor{red}{#1}}}}
\newcommand{\secondres}[1]{{\underline{\textcolor{blue}{#1}}}}

\usepackage[textsize=tiny]{todonotes}

\newcommand{\icon}{\raisebox{-2pt}{\includegraphics[width=1.0em]{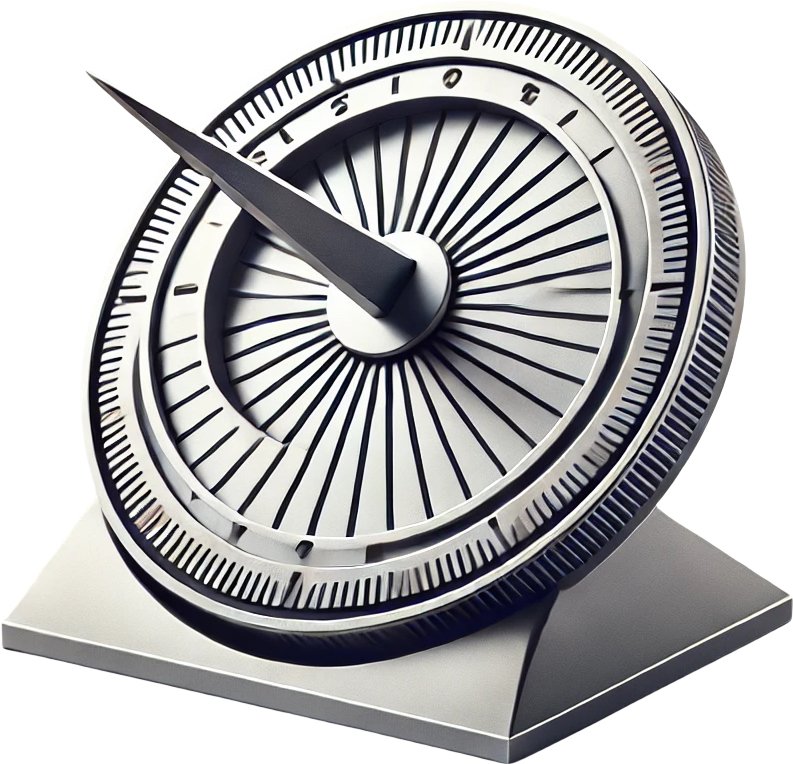}}\xspace}

\icmltitlerunning{Sundial: A Family of Highly Capable Time Series Foundation Models}

\begin{document}

\twocolumn[
\icmltitle{\icon Sundial: A Family of Highly Capable Time Series Foundation Models}



\icmlsetsymbol{equal}{*}

\begin{icmlauthorlist}
\icmlauthor{Yong Liu}{equal,software}
\icmlauthor{Guo Qin}{equal,software}
\icmlauthor{Zhiyuan Shi}{software}
\icmlauthor{Zhi Chen}{software}
\icmlauthor{Caiyin Yang}{software} \\
\icmlauthor{Xiangdong Huang}{software}
\icmlauthor{Jianmin Wang}{software}
\icmlauthor{Mingsheng Long}{software}
\end{icmlauthorlist}

\icmlaffiliation{software}{School of Software, BNRist, Tsinghua University. Yong Liu $<$liuyong21@mails.tsinghua.edu.cn$>$. Guo Qin $<$qinguo24@mails.tsinghua.edu.cn$>$}
\icmlcorrespondingauthor{Mingsheng Long}{mingsheng@tsinghua.edu.cn}

\icmlkeywords{time series, foundation models, pre-training, Transformers}

\vskip 0.3in
]



\printAffiliationsAndNotice{\icmlEqualContribution} % otherwise use the standard text.

\begin{abstract}
We introduce \emph{Sundial}, a family of native, flexible, and scalable time series foundation models. To predict the next-patch's distribution, we propose a \emph{TimeFlow Loss} based on flow-matching, which facilitates native pre-training of Transformers on continuous-valued time series without discrete tokenization. Conditioned on arbitrary-length time series, our models are pre-trained without specifying any prior distribution and can generate multiple probable predictions, achieving more flexibility in representation learning than using parametric densities. Towards time series foundation models, we leverage minimal but crucial adaptations of Transformers and curate \emph{TimeBench with one trillion time points}, comprising mostly real-world datasets and synthetic data. By mitigating mode collapse via TimeFlow Loss, we pre-train a family of Sundial models on TimeBench, which achieve unprecedented model capacity and generalization performance. In addition to excellent scalability, Sundial achieves state-of-the-art results on both point and probabilistic forecasting benchmarks with a just-in-time inference speed, i.e., making zero-shot predictions within a few milliseconds. We believe that Sundial's pioneering generative forecasting capability can improve model reliability in real-world decision-making. Code is available at: \href{https://github.com/thuml/Sundial}{https://github.com/thuml/Sundial}.
\end{abstract}

\section{Introduction}
Time series forecasting has fascinated people for thousands of years. Although people have been able to determine the time using instruments like sundials in 3000 BC, time series forecasting is intrinsically \emph{non-deterministic}~\cite{box2015time}. Therefore, generating a variety of probable predictions is crucial for decision-making. The growing demand has facilitated numerous statistical approaches over the past decades~\cite{hyndman2018forecasting, box2013box}, which provide high-profile theories and probabilistic tools for making reliable schedules. Recent advancements bring the boom of deftly designed models that automatically learn intricate dynamics and correlations from raw data~\cite{oreshkin2019n, nie2022time, zhang2023crossformer, liu2023itransformer}. Despite the impressive performance, deep models necessitate task-specific training on sufficient in-distribution data. Motivated by advances in large models~\cite{bommasani2021opportunities}, pre-trained time series foundation models have shown promising capabilities in out-of-distribution tasks~\cite{das2023decoder, liutimer, woo2024unified, ansari2024chronos}.

\begin{figure}[t]
\begin{center}
    \centerline{\includegraphics[width=\columnwidth]{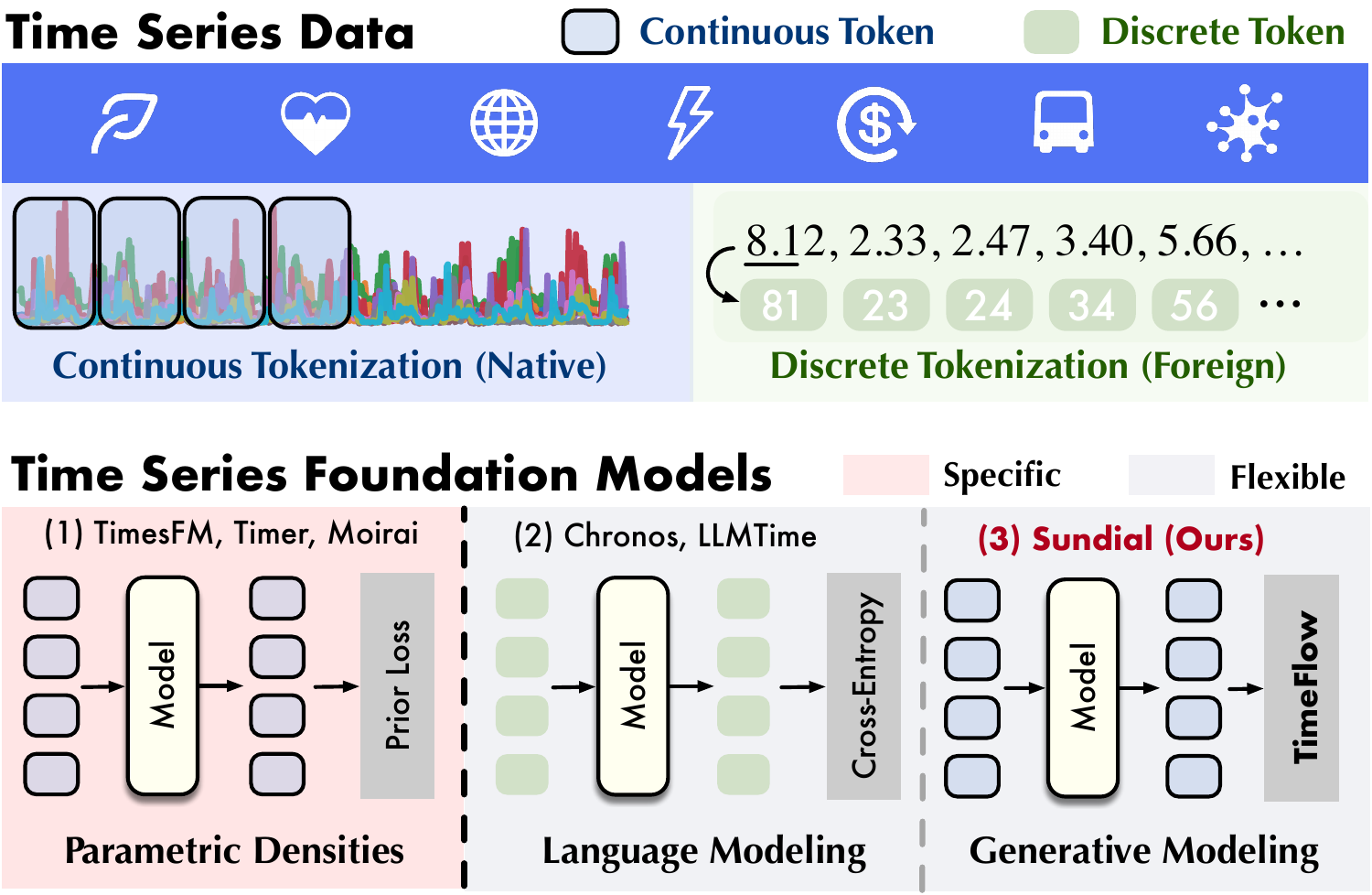}}
    \vspace{-8pt}
	\caption{A \emph{native} time series model operates on the original series of continuous values. A \emph{flexible} foundation model is pre-trained without specifying prior distributions. Sundial is the first family of native and flexible time series foundation models.}
	\label{fig:motivation}
\end{center}
\vspace{-30pt}
\end{figure}

Current research on time series foundation models has converged on building unified, scalable, and out-of-the-box forecasters, exhibiting zero-shot performance close to or sometimes surpassing supervised methods~\cite{aksugift}. Notably, Transformers~\cite{radford2018improving} are currently the \emph{de facto} architecture of these models. While pre-trained Transformers with an inherent generative ability have facilitated great success in language, image, and video generation~\cite{ramesh2021zero, openai2023gpt, liu2024sora}, most time series foundation models are not ``generative'' or, more specifically, probabilistic forecasters, thereby limiting reliability in decision-making. Although parametric densities specified with prior distributions~\cite{wen2017multi, woo2024unified} can be adopted to address uncertainty in time series forecasting, they can reduce the capacity of distributions learned by pre-trained models, especially on time series modality characterized by high heterogeneity. To learn arbitrarily intricate distributions without mode collapse, language modeling~\cite{bengio2000neural} that learns the categorical distribution via cross-entropy loss inspires subsequent works~\cite{gruver2023large, ansari2024chronos}, which treat time series as a \emph{foreign} language using discrete tokenization. Still, discrepancies between continuous-valued time series and discrete language tokens can lead to out-of-vocabulary issues and coarse-grained prediction intervals.

As shown in Figure~\ref{fig:motivation}, Sundial is presented as the first family of generative models among time series foundation models. As foundation models intend to learn complicated distributions from extensive datasets and facilitate transferability across agnostic downstream datasets, we do not specify any prior parametric densities, such as unimodal and multimodal Gaussian mixtures. Instead, we delve into generative modeling to tame Transformers as native, flexible, and scalable time series foundation models. By comparing to denoising diffusion models~\cite{li2024autoregressive}, we opt for a simple yet effective flow-matching framework~\cite{lipman2022flow}, which provides notable efficiency and sample quality~\cite{tong2023improving}. We propose \emph{TimeFlow Loss}, a parameterized training objective~\cite{zhang2018unreasonable} for autoregressive models to learn and sample from each token's predictive distribution. Optimizing models in the original continuous-valued domain, TimeFlow Loss facilitates patch-level generation and enables fast inference, which is naturally compatible with the time series modality.

In addition to TimeFlow, we enhance the Transformer with minimal but critical adaptations. We develop feasible patch tokenization for arbitrary-length input time series. We adopt RoPE~\cite{su2024roformer}, Pre-LN~\cite{xiong2020layer}, FlashAttention~\cite{dao2022flashattention}, and KV Cache~\cite{pope2023efficiently}, which are crucial but generally neglected in the development of time series foundation models. Besides, we pre-train our models by multi-patch prediction to reduce autoregression steps. We realize a rapid generation of multiple samples by reusing a shared lookback representation. Beyond facilitating scalable pre-training, these adaptations help real-time long-context inference and long-term generation.

To validate the scaling law of time series foundation models, we collect and curate \emph{TimeBench} with an unprecedented volume of a trillion time points. We present \emph{Sundial} as a family of highly capable foundation models, which achieve state-of-the-art on three large-scale and best-recognized benchmarks, including Time-Series-Library (TSLib)~\cite{wu2022timesnet}, GIFT-Eval~\cite{aksugift}, and FEV~\cite{ansari2024chronos}. Our contributions lie in these aspects:
\begin{itemize}
    \item We propose TimeFlow Loss to predict next-patch's distribution, allowing Transformers to be trained without discrete tokenization and make probable predictions.
    \item We present Sundial, a family of scalable and efficient time series foundation models built upon our enhanced Transformer and pre-trained on a trillion time points.
    \item Experimentally, Sundial achieves state-of-the-art zero-shot performance on point and probabilistic forecasting benchmarks, including TSLib, GIFT-Eval, and FEV, indicating a promising generative approach for the future improvement of time series foundation models.
\end{itemize}

\section{Related Work}
\subsection{Time Series Forecasting}
Forecasting is essential for decision-making. Advancements in deep learning for time series include theory-inspired deep modules~\cite{wu2021autoformer, liu2023koopa, wu2022timesnet}, architecture-oriented adaptations~\cite{bai2018empirical, salinas2020deepar, lim2021temporal}, and time series preprocessing~\cite{kim2021reversible, nie2022time}. Deep models learn the dataset-level distribution and benefit from strong generalization and model capacity. Statistical methods conduct case-by-case fitting on input series, achieving notable performance on small data~\cite{ke2017lightgbm, hyndman2018forecasting}.

One of the efforts towards more capable forecasters focuses on the foundation models~\cite{bommasani2021opportunities}, which address data-scarce scenarios by pre-training. More capable models support zero-shot forecasting, making inferences as fast as statistical methods and possessing large model capacity as deep models. Another aspect is to address uncertainty in time series forecasting. There is a growing research emphasis on probabilistic forecasting~\cite{woo2024unified, ansari2024chronos}. While parametric densities can be adopted as training objectives of probabilistic forecasting, they can be too specific to meet the heterogeneity of large-scale datasets, resulting in mode collapse in representation learning and over-smooth predictions (Figure~\ref{fig:showcases_compare1}-\ref{fig:showcases_compare2}). In this work, we introduce generative time series foundation models, which naturally address the uncertainty in forecasting.

\subsection{Time Series Foundation Models}
Recent research has concentrated on building versatile large time series models~\cite{liang2024foundation}. With the advances made in large language models, Transformer has become the dominant architecture. Several works adapt Transformers to address the unique 2D-dimensionality and heterogeneity of time series~\cite{woo2024unified, liu2024timer}. Specifically, our work delves into tokenization and optimization. Models such as TimesFM~\cite{das2023decoder}, Timer~\cite{liu2024timer, liutimer}, and Time-MoE~\cite{shi2024time} embed continuous values and fit unimodal distributions via MSE or quantile loss~\cite{wen2017multi}. However, prior loss may result in mode collapse because predictive distributions are highly divergent across different domains. Besides, these models cannot provide the confidence level of predictions, limiting reliability for decision-making. Based on continuous tokenization, Moirai~\cite{woo2024unified} presents a probabilistic model learning a mixture of distributions, but this prior can still fail to accommodate complex distributions. Inspired by language modeling, Chronos~\cite{ansari2024chronos} discretizes series via bucket quantization, learning more flexible categorical distributions by cross-entropy. Still, discrete tokenization is applied at each time point, which can lead to long contexts. Also, the final performance can be sensitive to quantization techniques. Unlike before, we tame Transformers as native time series foundation models, learning flexible distributions without discrete tokenization.

\subsection{Generative Modeling for Time Series}
By addressing complicated distributions during pre-training, generative modeling has become a focal point in the development of various foundation models~\cite{zhao2023survey, liu2024sora}. While this direction for time series mostly focused on time series generation~\cite{tashiro2021csdi} and task-specific forecasters~\cite{rasul2021autoregressive,shen2023non,kollovieh2024flow}, generative modeling for time series foundation models is hardly explored.
With the comparable flexibility in distribution learning as language modeling, diffusion denoising~\cite{sohl2015deep} and flow-matching~\cite{lipman2022flow} have gained increasing prevalence in continuous-valued modalities~\cite{lipman2024flow}. Compared with diffusion denoising models, flow-matching provides a simple yet efficient framework. With fewer steps involved in the forward and reverse processes, large models based on flow-matching have shown superior performance in image generation~\cite{esser2024scaling}.

Despite the connection in value continuity, generating images and future time series are fundamentally different tasks due to the autoregressive property of forecasting. Our proposed TimeFlow Loss is designed for autoregressive models to conduct conditional generation, which is a parameterized loss function~\cite{zhang2018unreasonable} for arbitrary distributions and enhances representation learning of foundation models.

\section{Preliminaries}

\subsection{Flow-Matching}
The goal of generative modeling is to learn the underlying probability distribution that generates the data. The framework of flow-matching transforms a sample $\mathbf{x}_0\sim p_0$ drawn from a source distribution into a sample $\mathbf{x}_1\sim p_1$ drawn from a target distribution. The transformation is continuous in time. For $d$-dimensional distributions, it is defined by a time-dependent velocity field $u_t : [0,1]\times\mathbb{R}^d\to\mathbb{R}^d$ , which is the solution of the ordinary differential equation (ODE):
\begin{equation*}
\frac{\mathrm{d}}{\mathrm{d} t} \psi_t(\mathbf{x})=u_t\big(\psi_t(\mathbf{x})\big)\ \text{and}\ \psi_0(\mathbf{x})=\mathbf{x}.
\end{equation*}
The velocity field $u_t$ determines a flow $\psi_t$. For all $t\in[0,1]$, $\psi_t$ generates the probability path $p_t$ that interpolates $p_0$ and $p_1$, i.e., $\mathbf{x}_t=\psi_t\left(\mathbf{x}_0\right) \sim p_t$ for $\mathbf{x}_0\sim p_0$. The implementation of flow-matching is to train a network $u^\theta_t$ parametrized by $\theta$ to fit the velocity field $u_t$, which is a regression-based task formulated as the Flow-Matching objective: 
\begin{equation*}
\mathcal{L}_{\mathrm{FM}}(\theta)=\mathbb{E}_{t,\mathbf{x}_t}\left\|u^\theta_t(\mathbf{x}_t)-u_t(\mathbf{x}_t)\right\|^2.
\end{equation*}
Furthermore, \citet{lipman2022flow} proved the equivalence of optimizing the Conditional Flow-Matching objective:
\begin{equation*}
\mathcal{L}_{\mathrm{CFM}}(\theta)=\mathbb{E}_{t,\mathbf{x}_t,\mathbf{x}_1}\left\|u^\theta_t(\mathbf{x}_t)-u_t(\mathbf{x}_t|\mathbf{x}_1)\right\|^2.
\end{equation*}
Leveraging the conditional optimal-transport (linear) path and a source Gaussian, the objective can be formulated as:
\begin{equation}\label{equ:cfm}
\mathcal{L}^{\mathrm{Gauss}}_{\mathrm{CFM}}(\theta)=\mathbb{E}_{t,\epsilon,\mathbf{x}_1 }\left\|u^\theta_t(\mathbf{x}_t)-(\mathbf{x}_1-\mathbf{x}_0)\right\|^2.
\end{equation}
where $t\sim\mathcal{U}[0,1], \mathbf{x}_0\sim\mathcal{N}(0,1)$ and $\mathbf{x}_t=t\mathbf{x}_1+(1-t)\epsilon$.

Consequently, we can train a generative network on given samples from the target distribution, and generate new samples by applying a push-forward process on samples drawn from a simple source Gaussian distribution:
\begin{equation}\label{equ:pf}
\mathbf{x}_{t+\Delta t}-\mathbf{x}_t=u^\theta_t(\mathbf{x}_t)\Delta t,\ \mathbf{x}_0\sim\mathcal{N}(\mathbf{0},\mathbf{I}),\ t\in[0, 1].
\end{equation}

\begin{figure*}[ht]
\begin{center}
    \center{\includegraphics[width=\textwidth]{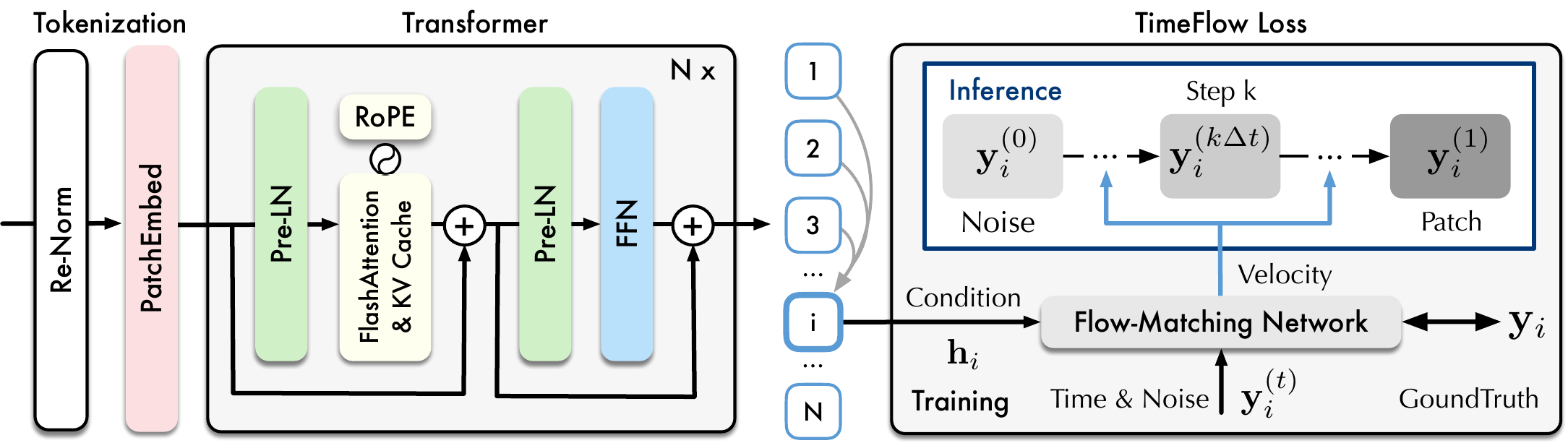}}
    \vspace{-20pt}
	\caption{Overall architecture of Sundial. The input time series is divided into patch tokens, which are embedded from original continuous values. The patch embeddings are fed into a decoder-only Transformer, a stable and speedup version that learns token representations via causal self-attention. The model is optimized using our TimeFlow Loss, a parameterized loss function that models per-token probability distribution conditioned on the learned representations, and generates multiple plausible predictions under the flow-matching framework.}
	\label{fig:architecture}
\end{center}
\vspace{-5pt}
\end{figure*}

\subsection{Generative Models for Probabilistic Forecasting}
Given a historical observation $x_{1:t}=\{x_1 , . . . , x_t\}$, the target of time series forecasting is to predict future time series $x_{t+1:t+f}=\{x_{t+1}, \dots, x_{t+f}\}$. The task can be generally formulated as $p(x_{t+1:t+f}|\mathbf{h}_{t})$, where $\mathbf{h}_{t}=f_\phi(x_{1:t})$ is the learned representation from a deep model $f_\phi$. In probabilistic forecasting, explicit optimization objectives are utilized to predict the statistics of future series, e.g., MSE or quantile loss, which have specified $p$ as a prior distribution. While using one parametric density generally fits well on a small amount of data, it can be the major bottleneck for scaling time series foundation models. Inspired by the success of large generative models~\cite{rombach2022high, openai2023gpt, esser2024scaling}, we introduce generative modeling to realize probabilistic forecasting:
\begin{equation}\label{equ:gf}
p_{\theta}(x_{t+1:t+f}|\mathbf{h}_{t})=g_{\theta}\big(f_\phi({x_{1:t}})\big).
\end{equation}
$g_\theta$ is a small trainable generative network conditioned on the learned representations of $f_\phi$, which is jointly optimized with $f_\phi$. While the generative model automatically fits the target distribution, it can sample raw predictions and calculating their statistics for probabilistic forecasting. The aim is conceptually related to conformal prediction~\cite{vovk2005algorithmic} but models uncertainty beyond prediction intervals. 

\section{Approach}
In this work, we conduct a univariate pre-training paradigm, which adopts the S3 format proposed by~\citet{liutimer} to address multivariate data. To mitigate value range discrepancy, we conduct normalization on time series individually per variable. Afterwards, we sample varying-length training samples with the maximum context length of $2880$. As a foundation model, Sundial is required to predict on out-of-distribution series with varied lengths during inference.

\subsection{Sundial}
As shown in Figure~\ref{fig:architecture}, the Sundial models consist of three parts: (1) time series tokenization, including a context-level re-normalization and a patch embedding that addresses any-length time series, (2) a Transformer backbone that learns the per-token representation of time series, and (3) \emph{TimeFlow Loss}, a parameterized loss function to model the per-token distribution and generate raw series during inference. 
Intuitively, Sundial can be regarded as an ARMA (Auto-Regression and Moving-Average) deep model, i.e., Transformer learns token representations autoregressively. Conditioned on the lookback representations, TimeFlow transforms random noises into non-deterministic predictions.

\subsubsection{Time Series Tokenization}

\paragraph{Re-Normalization} 
We adopt stationarization~\cite{liu2022non}, a non-parametric two-stage instance normalization conducted within each sample, which is initially proposed to mitigate non-stationarity of time series. Here, it helps to address temporal distribution shift and outlier ranges in input series, improving generalizability for zero-shot forecasting.

\paragraph{Patch Embedding} Given a univariate time series $\mathbf{X}=\{x_1, \dots, x_{T}\}$, it is divided into patches $\mathbf{x}_i=x_{1+(i-1)P:iP}$ with the length of $P$. To address non-divisible length, we pad the input at the beginning and use a binary mask $\mathbf{m}_i\in\mathbb{R}^P$ for each patch to indicate the padded position. It will lead to $N = \lceil T/P \rceil$ such input tokens. Subsequently, we use a shared MLP $: \mathbb{R}^{2P} \mapsto \mathbb{R}^D$ to embed all patch tokens:
\begin{equation}
    \mathbf{h}_i = \operatorname{PatchEmbed}\big(\operatorname{Concat}(\mathbf{x}_i, \mathbf{m}_i)\big),
\end{equation}
where $\mathbf{h}_i\in\mathbb{R}^D$ and $D$ is the dimension of token embedding. Unlike point-level quantization~\cite{ansari2024chronos}, we reserve original values without discrete quantization. It also reduces the context length (in token) of the Transformer. 
 
\subsubsection{Transformer Backbone}
Given $N$ token embeddings $\{\mathbf{h}_{i}\}$, we adopt several crucial adaptations on a decoder-only Transformer to obtain per-token representations aggregated from all previous tokens. First, we adapt Pre-LN~\cite{xiong2020layer} to improve pre-training stability. Second, we leverage a causal self-attention mechanism with RoPE~\cite{su2024roformer} that introduces the position information of patch tokens. It can be formulated as follows (the layer index is omitted for simplicity):
\begin{equation}
\begin{aligned}
\mathcal{A}_{ij} &= \mathbf{h}_{i}^\top\mathbf{W}_\mathbf{q}\mathbf{R}_{\Theta, i-j}\mathbf{W}_\mathbf{k}^\top\mathbf{h}_{j}, \\
\operatorname{Attention}(\mathbf{H}) &= \operatorname{Softmax}\left(\frac{\operatorname{Mask}(\mathcal{A})}{\sqrt{d}}\right) \mathbf{H}\mathbf{W}_\mathbf{v},
\end{aligned}
\end{equation}
where $\mathbf{W}_\mathbf{q}, \mathbf{W}_\mathbf{k}, \mathbf{W}_\mathbf{v} \in \mathbb{R}^{D\times d}$ project token embeddings $\mathbf{H}=\{\mathbf{h}_{i}\}$ into $d$-dimensional queries, keys, and values. $\mathbf{R}_{\Theta, t}\in\mathbb{R}^{d\times d}$ is the rotary matrix with rotation degree $(t \cdot \Theta)$. Lastly, we implement FlashAttention~\cite{dao2022flashattention} and KV Cache~\cite{pope2023efficiently}, since these enhancements for deployment are increasingly emphasized in large foundation models~\cite{shoeybi2019megatron, rasley2020deepspeed}.

\subsubsection{TimeFlow Loss} Given representations $\{\mathbf{h}_i\}$ extracted by the last layer of the Transformer, we aim to generate length-$F$ predictions $\widehat{\mathbf{y}}_i=\widehat{x}_{1+iP, F+iP}$ at each position $i$ via our autoregressive model. Motivated by the empirical observation that a larger patch size improves the performance in decoder-only Transformers~\cite{das2023decoder} while a small patch size can be more flexible to accommodate data of different frequencies, we adopt multi-patch predictions ($F>P$) for pre-training, which also reduces the steps of autoregressive inference.

Based on Equations~\ref{equ:cfm} and~\ref{equ:gf}, we formulate a new generative forecasting conditioned on a sequential representation $\mathbf{h}_i$:
\begin{equation}
\mathcal{L}(\theta, \mathbf{h}_i)=\mathbb{E}_{t,\epsilon,\mathbf{y}_i }\left\|u^\theta_t\big(\mathbf{y}_i^{(t)}|\mathbf{h}_i\big)-\big(\mathbf{y}_i-\mathbf{y}_i^{(0)}\big)\right\|^2.
\end{equation}
where $\mathbf{y}_i \in\mathbb{R}^F$ is the groundtruth value and $\mathbf{y}_i^{(0)}$ is a $F$-dimensional Gaussian noise, $t$ is sampled from $\mathcal{U}[0,1]$, and $\mathbf{y}_i^{(t)}=t\mathbf{y}_i+(1-t)\mathbf{y}_i^{(0)}$ is constructed by the conditional optimal-transport path. It is important to note that the conditional representation $\mathbf{h}_i$ differs from the conditional path and the conditional source distribution.
Instead, $\mathbf{h}_i$ is a condition of position $i$, also a time-invariant condition of the whole flow-matching process $t\in[0,1]$. Technically, we implement the flow-matching network by a small MLP:
\begin{equation}
u^\theta_t\big(\mathbf{y}_i^{(t)}|\mathbf{h}_i\big)=\operatorname{FM-Net}\big(\mathbf{y}_i^{(t)}, t, \mathbf{h}_i\big). 
\end{equation}
The training process involves sampling the noised $\mathbf{y}_i^{(t)}$, and jointly input it with $t$. The condition $\mathbf{h}_i$ is integrated into the flow-matching network via AdaLN~\cite{peebles2023scalable}. TimeFlow Loss for autoregressive models is formulated as:
\begin{equation}
\mathcal{L}_{\mathrm{TimeFlow}}=\sum_{i=1}^{N}\left\|\operatorname{FM-Net}\big(\mathbf{y}_i^{(t)}, t, \mathbf{h}_i\big)-\big(\mathbf{y}_i-\mathbf{y}_i^{(0)}\big)\right\|^2.
\end{equation}
\paragraph{Inference} Based on Equation~\ref{equ:pf}, the push-forward process conditioned on a learned representation $\mathbf{h}_i$ is formulated as
\begin{equation}
\mathbf{y}_i^{(t+\Delta t)} = \mathbf{y}_i^{(t)} + u^\theta_t\big(\mathbf{y}_i^{(t)}|\mathbf{h}_i\big)\Delta t.
\end{equation}
Technically, we adopt a $K$-step uniform trajectory, and set $\Delta t=1/K$. The sampling is done via starting from an initial Gaussian noise and advancing with the velocity generated by the trained $\operatorname{FM-Net}$ iteratively, as shown in Algorithm~\ref{alg:tf_sample}.

\begin{algorithm}
\caption{TimeFlow Loss: Sampling}
\begin{algorithmic}[1]\label{alg:tf_sample}
\REQUIRE condition $\mathbf{h}_i \in \mathbb{R}^{D}$, path steps $K$.
\STATE Sample initial noise $\widehat{\mathbf{y}}_i \sim \mathcal{N}(\mathbf{0}, \mathbf{I})$.
\STATE $\Delta t = 1/K$ 
\STATE $\textbf{for}\ k\ \textbf{in}\ \{0, 1 \dots, K-1\}\ \textbf{do}$
\STATE $\textbf{\textcolor{white}{for}}$\ $\widehat{\mathbf{y}}_i \leftarrow \widehat{\mathbf{y}}_i + \operatorname{FM-Net}\big(\widehat{\mathbf{y}}_i, k\Delta t, \mathbf{h}_i\big)\Delta t$
\STATE \textbf{end for}
\STATE \textbf{Return:} $\widehat{\mathbf{y}}_i$
\end{algorithmic}
\end{algorithm}

This procedure generates a predicted sample $\widehat{\mathbf{y}}_i$ at position $i$. To calibrate probabilistic forecasting results during inference, we repeat this procedure using different initial noises and estimate statistics such as the median and quantiles from a set of generated predictions. We implement an efficient repeated-sampling in the TimeFlow module. The condition (representation) of lookback series is shared and reused for different initial noises, thereby reducing the overhead of repeated forwarding in the Transformer backbone.

\subsection{TimeBench}

\begin{figure}[ht]
\begin{center}
    \centerline{\includegraphics[width=\columnwidth]{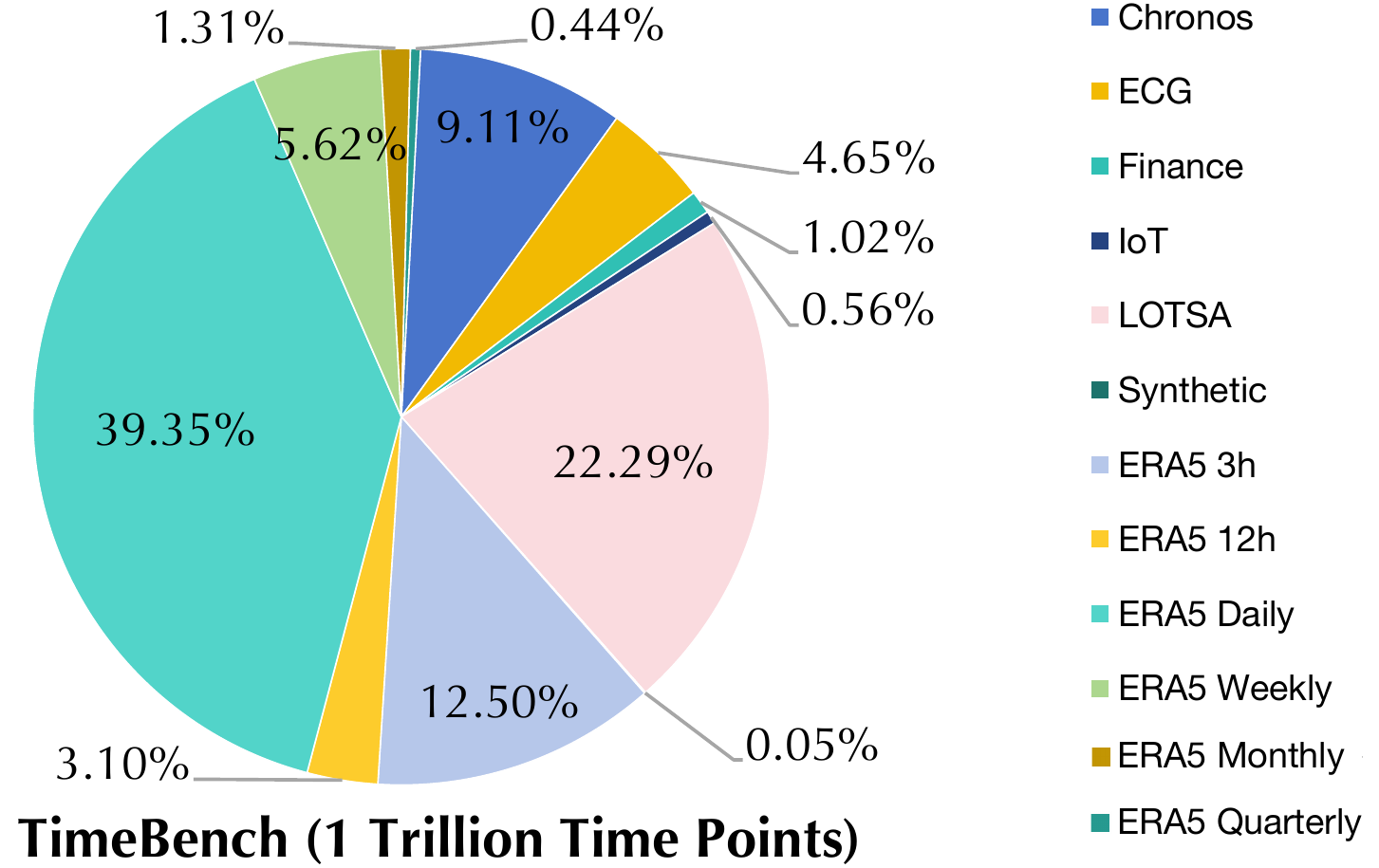}}
    \vspace{-8pt}
	\caption{Ratios of data sources in TimeBench, the pre-training corpora of Sundial. Detailed statistics are provide in Table~\ref{tab:dataset_summary}.}
	\label{fig:timebench}
\end{center}
\vspace{-25pt}
\end{figure}

We collected and curated \emph{TimeBench}, which comprises over a trillion time points from various sources, as shown in Figure~\ref{fig:timebench}. Several datasets originate from research teams~\cite{woo2024unified, ansari2024chronos, liu2024timer, liutimer}. While most datasets are collected from real-world records, a small portion (0.05\%) is generated synthetically to enhance pattern diversity, following KernelSynth proposed by~\citet{ansari2024chronos}. We also leverage substantial meteorological data~\cite{hersbach2020era5} because of the predictability of weather systems. Data of different frequencies encompasses common and comprehensive temporal dynamics.

\section{Experiments}
We evaluate Sundial on best-recognized zero-shot forecasting benchmarks (Section~\ref{sec:zsf}) and investigate the scaling behavior of Sundial (Section~\ref{sec:scale}). We compare TimeFlow with other training objectives (Section~\ref{sec:timeflow}). We delve into test-time calibration of generative forecasters (Section~\ref{sec:inference}). We conduct model adaptation of Sundial, i.e., instruction tuning (Section~\ref{sec:adapt}) and provide in-depth ablation studies to evaluate our modular enhancement (Section~\ref{sec:ablation}).

\begin{table*}[ht]
  \caption{Zero-shot forecasting results of time series foundation models on long-term forecasting datasets (Time-Series-Library)~\cite{wu2022timesnet}. Corresponding prediction lengths include $\{96, 192, 336, 720\}$. A lower MSE or MAE indicates a better prediction. Averaged results of four prediction lengths are reported here. $1^{\text{st}}$ Count represents the number of wins achieved by a model under all prediction lengths and datasets. Results of baseline models are officially reported by \citet{shi2024time}. Datasets in pre-training are not evaluated on corresponding models, which are denoted by the dash ($-$). Full results under all prediction lengths are provided in Table~\ref{tab:zero_shot_datasets_full}.}
  \vspace{-5pt}
  \renewcommand{\arraystretch}{0.85} 
  \centering
  \begin{threeparttable}
  \begin{small}
  \renewcommand{\multirowsetup}{\centering}
  \setlength{\tabcolsep}{1.1pt}
  \label{tab:zero_shot_datasets}
  \begin{tabular}{c|c|cc|cc|cc|cc|cc|cc|cc|cc|cc|cc|cc|cc}
    \toprule
    \multicolumn{2}{c}{\multirow{2}{*}{Models}} & 
    \multicolumn{2}{c}{\rotatebox{0}{\scalebox{0.8}{$\textbf{Sundial}_{\textit{Small}}$}}} &
    \multicolumn{2}{c}{\rotatebox{0}{\scalebox{0.8}{$\textbf{Sundial}_{\textit{Base}}$}}} &
    \multicolumn{2}{c}{\rotatebox{0}{\scalebox{0.8}{$\textbf{Sundial}_{\textit{Large}}$}}} &

    \multicolumn{2}{c}{\rotatebox{0}{\scalebox{0.7}{$\textbf{Time-MoE}_{\textit{Base}}$}}} &
    \multicolumn{2}{c}{\rotatebox{0}{\scalebox{0.7}{$\textbf{Time-MoE}_{\textit{Large}}$}}} &
    \multicolumn{2}{c}{\rotatebox{0}{\scalebox{0.7}{$\textbf{Time-MoE}_{\textit{Ultra}}$}}}  &
        \multicolumn{2}{c}{\rotatebox{0}{\scalebox{0.8}{$\textbf{Timer-XL}$}}} &
    \multicolumn{2}{c}{\rotatebox{0}{\scalebox{0.8}{{$\textbf{Moirai}_{\textit{Base}}$}}}} &
    \multicolumn{2}{c}{\rotatebox{0}{\scalebox{0.8}{$\textbf{Moirai}_{\textit{Large}}$}}}&
    \multicolumn{2}{c}{\rotatebox{0}{\scalebox{0.8}{$\textbf{Chronos}_{\textit{Base}}$}}} &
    \multicolumn{2}{c}{\rotatebox{0}{\scalebox{0.8}{$\textbf{Chronos}_{\textit{Large}}$}}} &
    \multicolumn{2}{c}{\rotatebox{0}{\scalebox{0.8}{$\textbf{TimesFM}$}}}\\
    \multicolumn{2}{c}{} &
    \multicolumn{2}{c}{\scalebox{0.8}{\textbf{(Ours)}}} & 
    \multicolumn{2}{c}{\scalebox{0.8}{\textbf{(Ours)}}} & 
    \multicolumn{2}{c}{\scalebox{0.8}{\textbf{(Ours)}}} & 

    \multicolumn{2}{c}{\scalebox{0.8}{\citeyearpar{shi2024time}}} & 
    \multicolumn{2}{c}{\scalebox{0.8}{\citeyearpar{shi2024time}}} & 
    \multicolumn{2}{c}{\scalebox{0.8}{\citeyearpar{shi2024time}}}  & 
        \multicolumn{2}{c}{\scalebox{0.8}{\citeyearpar{liu2024timer}}} & 
    \multicolumn{2}{c}{\scalebox{0.8}{\citeyearpar{woo2024unified}}} & 
    \multicolumn{2}{c}{\scalebox{0.8}{\citeyearpar{woo2024unified}}}& 
    \multicolumn{2}{c}{\scalebox{0.8}{\citeyearpar{ansari2024chronos}}} &
    \multicolumn{2}{c}{\scalebox{0.8}{\citeyearpar{ansari2024chronos}}} &
    \multicolumn{2}{c}{\scalebox{0.8}{\citeyearpar{das2023decoder}}} \\
    \cmidrule(lr){3-4} \cmidrule(lr){5-6}\cmidrule(lr){7-8} \cmidrule(lr){9-10}\cmidrule(lr){11-12}\cmidrule(lr){13-14} \cmidrule(lr){15-16} \cmidrule(lr){17-18} \cmidrule(lr){19-20} \cmidrule(lr){21-22} \cmidrule(lr){23-24} \cmidrule(lr){25-26}
    \multicolumn{2}{c}{Metric}  & \scalebox{0.78}{MSE} & \scalebox{0.78}{MAE}  & \scalebox{0.78}{MSE} & \scalebox{0.78}{MAE}  & \scalebox{0.78}{MSE} & \scalebox{0.78}{MAE}  & \scalebox{0.78}{MSE} & \scalebox{0.78}{MAE}  & \scalebox{0.78}{MSE} & \scalebox{0.78}{MAE}  & \scalebox{0.78}{MSE} & \scalebox{0.78}{MAE}  & \scalebox{0.78}{MSE} & \scalebox{0.78}{MAE} & \scalebox{0.78}{MSE} & \scalebox{0.78}{MAE} & \scalebox{0.78}{MSE} & \scalebox{0.78}{MAE} & \scalebox{0.78}{MSE} & \scalebox{0.78}{MAE} & \scalebox{0.78}{MSE} & \scalebox{0.78}{MAE} & \scalebox{0.78}{MSE} & \scalebox{0.78}{MAE} \\
    \toprule
    \multicolumn{2}{c|}{\scalebox{0.78}{ETTm1}}
     &\scalebox{0.78}{0.354} 
    &\scalebox{0.78}{0.388}
    &\secondres{\scalebox{0.78}{0.336}} 
    &\secondres{\scalebox{0.78}{0.377}}
    &\boldres{\scalebox{0.78}{0.331}} 
    &\boldres{\scalebox{0.78}{0.369}}

    & \scalebox{0.78}{0.394} 
    & \scalebox{0.78}{0.415} 
    & \scalebox{0.78}{0.376} 
    & \scalebox{0.78}{0.405} 
    &\scalebox{0.78}{0.356}
    &\scalebox{0.78}{0.391}
        &\scalebox{0.78}{0.373}
    & \scalebox{0.78}{0.392} 
    & \scalebox{0.78}{0.406} 
    &\scalebox{0.78}{0.385}
    &{\scalebox{0.78}{0.422}} 
    &{\scalebox{0.78}{0.391}}
    &\scalebox{0.78}{0.645} 
    &\scalebox{0.78}{0.500} 
    &\scalebox{0.78}{0.555} 
    &\scalebox{0.78}{0.465} 
    &\scalebox{0.78}{0.433} 
    &\scalebox{0.78}{0.418} \\
    
    \midrule
    \multicolumn{2}{c|}{\scalebox{0.78}{ETTm2}}
    &\scalebox{0.78}{0.265} 
    &\scalebox{0.78}{0.324} 
    & \secondres{\scalebox{0.78}{0.258}} 
    &\secondres{\scalebox{0.78}{0.320}} 
    & \boldres{\scalebox{0.78}{0.254}} 
    &\boldres{\scalebox{0.78}{0.315}}

    & \scalebox{0.78}{0.317} 
    & \scalebox{0.78}{0.365} 
    & \scalebox{0.78}{0.316} 
    & \scalebox{0.78}{0.361} 
    & {\scalebox{0.78}{0.288}} 
    & \scalebox{0.78}{0.344} 
        & {{\scalebox{0.78}{0.273}}} 
    & {{\scalebox{0.78}{0.336}}} 
    & {\scalebox{0.78}{0.311}}
    &{{\scalebox{0.78}{0.337}}} 
    &\scalebox{0.78}{0.329} 
    &\scalebox{0.78}{0.343} 
    & \scalebox{0.78}{0.310} 
    &\scalebox{0.78}{0.350} 
    &\scalebox{0.78}{0.295} 
    &\scalebox{0.78}{0.338} 
    &\scalebox{0.78}{0.328} 
    &\scalebox{0.78}{0.346}\\
    
    \midrule
    \multicolumn{2}{c|}{\scalebox{0.78}{ETTh1}}
    & \boldres{\scalebox{0.78}{0.390}} 
     & \secondres{\scalebox{0.78}{0.418}}
     & \scalebox{0.78}{0.411} 
     & \scalebox{0.78}{0.434} 
     & \scalebox{0.78}{0.395} 
     & \scalebox{0.78}{0.420} 

     & \scalebox{0.78}{0.400} 
     & \scalebox{0.78}{0.424} 
     & \secondres{\scalebox{0.78}{0.394}} 
     & \scalebox{0.78}{0.419} 
     & \scalebox{0.78}{0.412} 
     & \scalebox{0.78}{0.426} 
          & \scalebox{0.78}{0.404} 
     & \boldres{\scalebox{0.78}{0.417}}
     & \scalebox{0.78}{0.417} 
     & \scalebox{0.78}{0.419} 
     & \scalebox{0.78}{0.480} 
     & \scalebox{0.78}{0.439} 
     & \scalebox{0.78}{0.591} 
     & \scalebox{0.78}{0.468} 
     & \scalebox{0.78}{0.588} 
     & \scalebox{0.78}{0.466} 
     & \scalebox{0.78}{0.473} 
     & \scalebox{0.78}{0.443}\\
     
    \midrule
    \multicolumn{2}{c|}{\scalebox{0.78}{ETTh2}}
    & {\scalebox{0.78}{0.340}} 
     & \scalebox{0.78}{0.387} 
     & \boldres{\scalebox{0.78}{0.333}} 
     & \scalebox{0.78}{0.387} 
     & \secondres{\scalebox{0.78}{0.334}}
     & \scalebox{0.78}{0.387} 

     & \scalebox{0.78}{0.366} 
     & \scalebox{0.78}{0.404} 
     & \scalebox{0.78}{0.405} 
     & \scalebox{0.78}{0.415} 
     & \scalebox{0.78}{0.371} 
     & \scalebox{0.78}{0.399} 
          & \scalebox{0.78}{0.347} 
     & \scalebox{0.78}{0.388} 
     & \scalebox{0.78}{0.362} 
     & \secondres{\scalebox{0.78}{0.382}}
     & \scalebox{0.78}{0.367} 
     & \boldres{\scalebox{0.78}{0.377}}
     & \scalebox{0.78}{0.405} 
     & \scalebox{0.78}{0.410} 
     & \scalebox{0.78}{0.455} 
     & \scalebox{0.78}{0.427}
     & \scalebox{0.78}{0.392}
     & \scalebox{0.78}{0.406} \\

    \midrule
    \multicolumn{2}{c|}{\scalebox{0.78}{ECL}}
     & \secondres{\scalebox{0.78}{0.169}} 
     & \secondres{\scalebox{0.78}{0.265}}
     &\secondres{\scalebox{0.78}{0.169}} 
     & \secondres{\scalebox{0.78}{0.265}}
     & \boldres{\scalebox{0.78}{0.166}} 
     & \boldres{\scalebox{0.78}{0.262}}

     & \scalebox{0.78}{-} 
     & \scalebox{0.78}{-} 
     & \scalebox{0.78}{-} 
     & \scalebox{0.78}{-} 
     & \scalebox{0.78}{-} 
     & \scalebox{0.78}{-} 
          & \scalebox{0.78}{0.174} 
     & \scalebox{0.78}{0.278} 
     & \scalebox{0.78}{0.187} 
     & \scalebox{0.78}{0.274} 
     & \scalebox{0.78}{0.186} 
     & \scalebox{0.78}{0.270} 
     & \scalebox{0.78}{0.214} 
     & \scalebox{0.78}{0.278} 
     & \scalebox{0.78}{0.204} 
     & \scalebox{0.78}{0.273} 
     & \scalebox{0.78}{-} 
     & \scalebox{0.78}{-}\\

    \midrule
    \multicolumn{2}{c|}{\scalebox{0.78}{Weather}}
     & \boldres{\scalebox{0.78}{0.233}} 
     & \secondres{\scalebox{0.78}{0.271}}
     & \secondres{\scalebox{0.78}{0.234}}
     & \boldres{\scalebox{0.78}{0.270}}
     & \scalebox{0.78}{0.238} 
     & \scalebox{0.78}{0.275}

     & \scalebox{0.78}{0.265}
     & \scalebox{0.78}{0.297} 
     & \scalebox{0.78}{0.270} 
     & \scalebox{0.78}{0.300} 
     & \scalebox{0.78}{0.256}
     & \scalebox{0.78}{0.288}
          & \scalebox{0.78}{0.256} 
     & \scalebox{0.78}{0.294}
     & \scalebox{0.78}{0.287}
     & \scalebox{0.78}{0.281} 
     & \scalebox{0.78}{0.264}
     & \scalebox{0.78}{0.273}
     & \scalebox{0.78}{0.292} 
     & \scalebox{0.78}{0.315} 
     & \scalebox{0.78}{0.279} 
     & \scalebox{0.78}{0.306} 
     & \scalebox{0.78}{-} 
     & \scalebox{0.78}{-}\\
     \midrule
    
    \multicolumn{2}{c|}{\textbf{\scalebox{0.78}{{$1^{\text{st}}$ Count}}}} &\scalebox{0.78}{7} 
     &\scalebox{0.78}{2} 
     & \secondres{\scalebox{0.78}{{8}}}
     & \scalebox{0.78}{{5}} 
     & \boldres{\scalebox{0.78}{16}}
     & \boldres{\scalebox{0.78}{{16}}}

     &\scalebox{0.78}{{0}} 
     &\scalebox{0.78}{{1}} 
     & \scalebox{0.78}{0} 
     & \scalebox{0.78}{0} 
     & \scalebox{0.78}{2} 
     & \scalebox{0.78}{1} 
          & \scalebox{0.78}{{1}} 
     & \scalebox{0.78}{3}
     & \scalebox{0.78}{0} 
     & \scalebox{0.78}{{2}} 
     & \scalebox{0.78}{0} 
     & \secondres{\scalebox{0.78}{6}}
     & \scalebox{0.78}{0} 
     & \scalebox{0.78}{0} 
     & \scalebox{0.78}{0} 
     & \scalebox{0.78}{0} 
     & \scalebox{0.78}{0} 
     & \scalebox{0.78}{0} \\ 
    \bottomrule
  \end{tabular}
    \end{small}
  \end{threeparttable}
\vspace{-10pt}
\end{table*}

\begin{table*}[hbtp]
    \caption{GIFT-Eval comprises $23$ datasets characterized by a variety of frequencies, variate numbers, and prediction lengths. We evaluate zero-shot performance using $100$ generated series, being consistent with~\citet{woo2024unified}. A lower MASE or CRPS indicates a better performance. Rank assigns a numerical ranking of all $97$ configurations. Baseline results are officially reported by~\citet{aksugift}.}
    \label{tab:gift_eval}
    \centering
    \vskip 0.05in
    \renewcommand{\multirowsetup}{\centering}
    \setlength{\tabcolsep}{5.1pt}
    \resizebox{\linewidth}{!}{
    \begin{tabular}{l|cccccccccccccc}
    \toprule
    \textbf{Type} & \multicolumn{4}{c}{\textbf{Statistical Methods}} & \multicolumn{5}{c}{\textbf{Task-Specific Models (Superwised)}} & \multicolumn{5}{c}{\textbf{Time Series Foundation Models (Zero-Shot)}} \\
    \cmidrule(lr){1-1} \cmidrule(lr){2-5} \cmidrule(lr){6-10} \cmidrule(lr){11-15}
        \multirow{2}{*}{\textbf{Model}}
            & \multirow{2}{*}{{Na\"ive}} & {Seasonal} & {Auto} & {Auto} & {DeepAR} & {TiDE} & {N-BEATS} & {PTST.} & {iTrans.} & {TimesFM} & {TabPFN} & {Chronos} & {Moirai} & \textbf{Sundial} \\
            & & {Na\"ive} & {ARIMA} & {Theta} & \citeyearpar{salinas2020deepar} & \citeyearpar{das2023long} & \citeyearpar{oreshkin2019n} & \citeyearpar{nie2022time} & \citeyearpar{liu2023itransformer} & \citeyearpar{das2023decoder} & \citeyearpar{hoo2025tabular} & \citeyearpar{ansari2024chronos} & \citeyearpar{woo2024unified} &  \textbf{(Ours)} \\
        \midrule
            \textbf{MASE} & 1.260 & 1.000 & 0.964 & 0.978 & 1.206  & 0.980 & 0.842 & 0.762 & 0.802 & \secondres{0.680} & 0.748 & 0.786 & 0.809 &  \boldres{0.673}\\
            \textbf{CRPS}  & 1.383 & 1.000 & 0.770 & 1.051 & 0.721 & 0.652 & 0.689 & 0.496 & 0.524 & \boldres{0.465} & 0.480 & 0.551 & 0.515 &  \secondres{0.472}\\
            \textbf{Rank}  & 28.072  & 26.175 & 21.515 & 24.031 & 18.938 & 18.557 & 21.381 & 10.052 & 11.320 & \boldres{8.237} & \secondres{8.268} & 14.309 & 10.175 & 9.062 \\
        \bottomrule
    \end{tabular}}
    \vspace{-10pt}
\end{table*}

\subsection{Time Series Forecasting}\label{sec:zsf}
In this section, we focus on zero-shot forecasting, we compare Sundial with advanced time series foundation models on various benchmarks, including (1) point forecasting: we adopt the long-term forecasting benchmark~\cite{wu2022timesnet}, which assesses the performance under different forecasting horizons using MSE and MAE; (2) probabilistic forecasting: we experiment on GIFT-Eval~\cite{aksugift} and FEV leaderboard~\cite{ansari2024chronos}, following their official evaluation suite and assessing point (MASE) and probabilistic (CRPS and WQL) metrics. All evaluated datasets are excluded from the pre-training dataset. Model is available on HuggingFace\footnote{\href{https://huggingface.co/thuml/sundial-base-128m}{https://huggingface.co/thuml/sundial-base-128m}.} and configurations are detailed in Table~\ref{tab:configuration}. 

\subsubsection{Point Forecasting}
As shown in Table~\ref{tab:zero_shot_datasets}, Sundial consistently outperforms other advanced time series foundation models. Compared with the previous state-of-the-art model Time-MoE~\cite{shi2024time}, the Sundial family using fewer parameters achieves the average MSE reduction of $7.57\%$ and averaged MAE reduction of $4.71\%$. Notably, continuous tokenization allows our model to conduct patch-level forecasting with fewer autoregression steps, while Chronos using point-wise discrete tokenization may not be suitable in long-term forecasting.

\begin{figure*}[ht]
\begin{center}
    \center{\includegraphics[width=\textwidth]{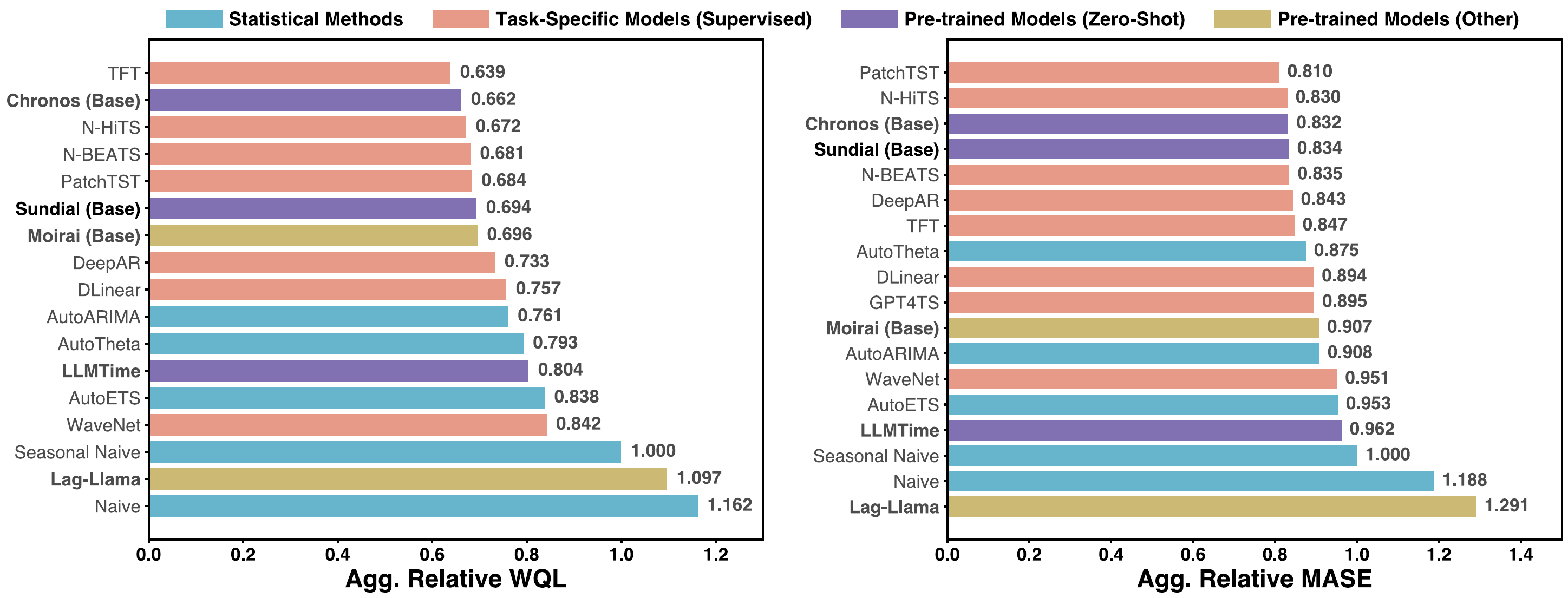}}
    \vspace{-20pt}
\caption{Model evaluation on the FEV leaderboard, which includes $27$ datasets not seen by Sundial. Baseline models can be categorized into statistical methods fitting on each time series, task-specific deep models trained on each dataset, and pre-trained foundation models. Pre-trained Models that have seen several datasets during pre-training are denoted as Pre-trained Models (Other). A lower MASE/WQL indicates a better result. Sundial makes probabilistic predictions using $20$ generated series, being consistent with~\citet{ansari2024chronos}.}
	\label{fig:fev}
\end{center}
\vspace{-10pt}
\end{figure*}

\subsubsection{Probabilistic Forecasting}
Beyond point forecasting, Sundial possesses a unique generative capability for making probabilistic predictions. Following~\citet{ansari2024chronos}, we calculate the median and quantiles using a set of raw predictions of Sundial. While several baseline models have been pre-trained by the consistent objective function for probabilistic evaluation, e.g., quantile loss for WQL, Sundial calculates these statistics for evaluation without any prior knowledge. 

\paragraph{GIFT-Eval} Aggregated results are presented in Table \ref{tab:gift_eval}. The benchmark evaluates performance from $23$ datasets and $13$ baseline models, encompassing statistical methods, task-specific models, and time series foundation models. Among supervised models and advanced foundation models, Sundial attains the first place in MASE and second place in CRPS on all unseen datasets. While the top PatchTST~\cite{nie2022time} is exhaustively trained and tweaked on each dataset, the zero-shot performance of Sundial highlights its simplicity and robustness on this comprehensive benchmark.

\begin{figure}[ht]
\begin{center}
    \centerline{\includegraphics[width=\columnwidth]{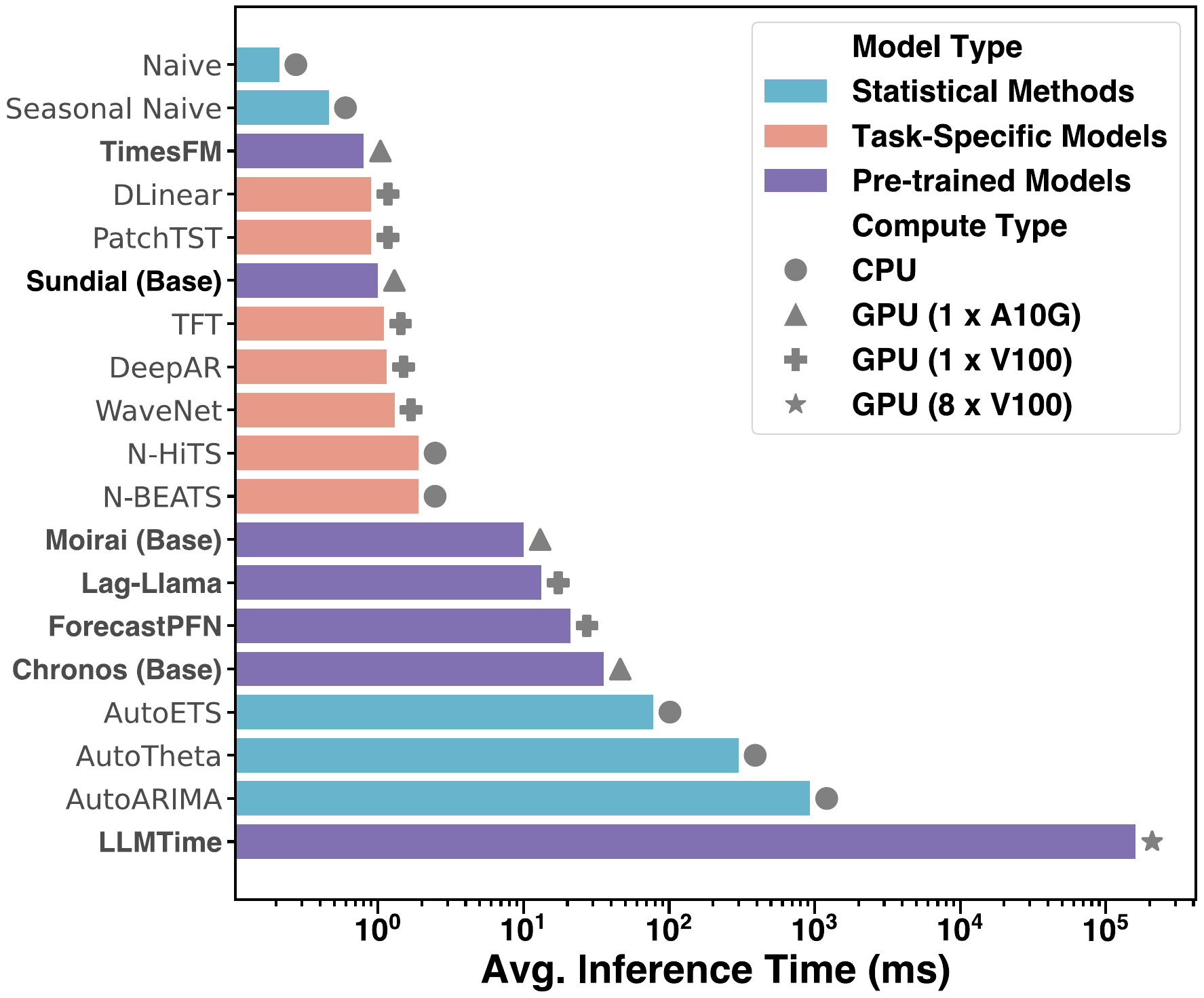}}
    \vspace{-5pt}
	\caption{Inference time evaluation following~\citet{ansari2024chronos}, which is averaged from the FEV leaderboard. Computing resources of different models are marked. We plot the logarithmic x-axis.}
	\label{fig:inference_time}
\end{center}
\vspace{-20pt}
\end{figure}

\paragraph{FEV Leaderboard} We evaluate our Sundial on the open leaderboard established by AutoGluon~\cite{ansari2024chronos}, which includes $27$ datasets for probabilistic forecasting. As shown in Figure~\ref{fig:fev}, the zero-shot forecasting performance of Sundial exceeds $70\%$ statistical methods and deep models that are superwisedly trained in distribution. While Sundial is ranked as the second zero-shot pre-trained models after Chronos, Sundial realizes $35\times$ inference speedup as shown in Figure~\ref{fig:inference_time}. Based on patch-wise tokenization and multi-patch prediction, our inference speed is near to N-BEATS.

Besides, we provide qualitative showcases in Appendix~\ref{app:showcase}. TimeFlow can generate highly eventful and coherent temporal patterns with input series. Beyond the mean or quantiles, our model enables the estimation of arbitrary statistics by sampling directly from the predictive distribution.

\subsection{Scalability}\label{sec:scale}
\begin{figure}[ht]
\begin{center}
    \centerline{\includegraphics[width=\columnwidth]{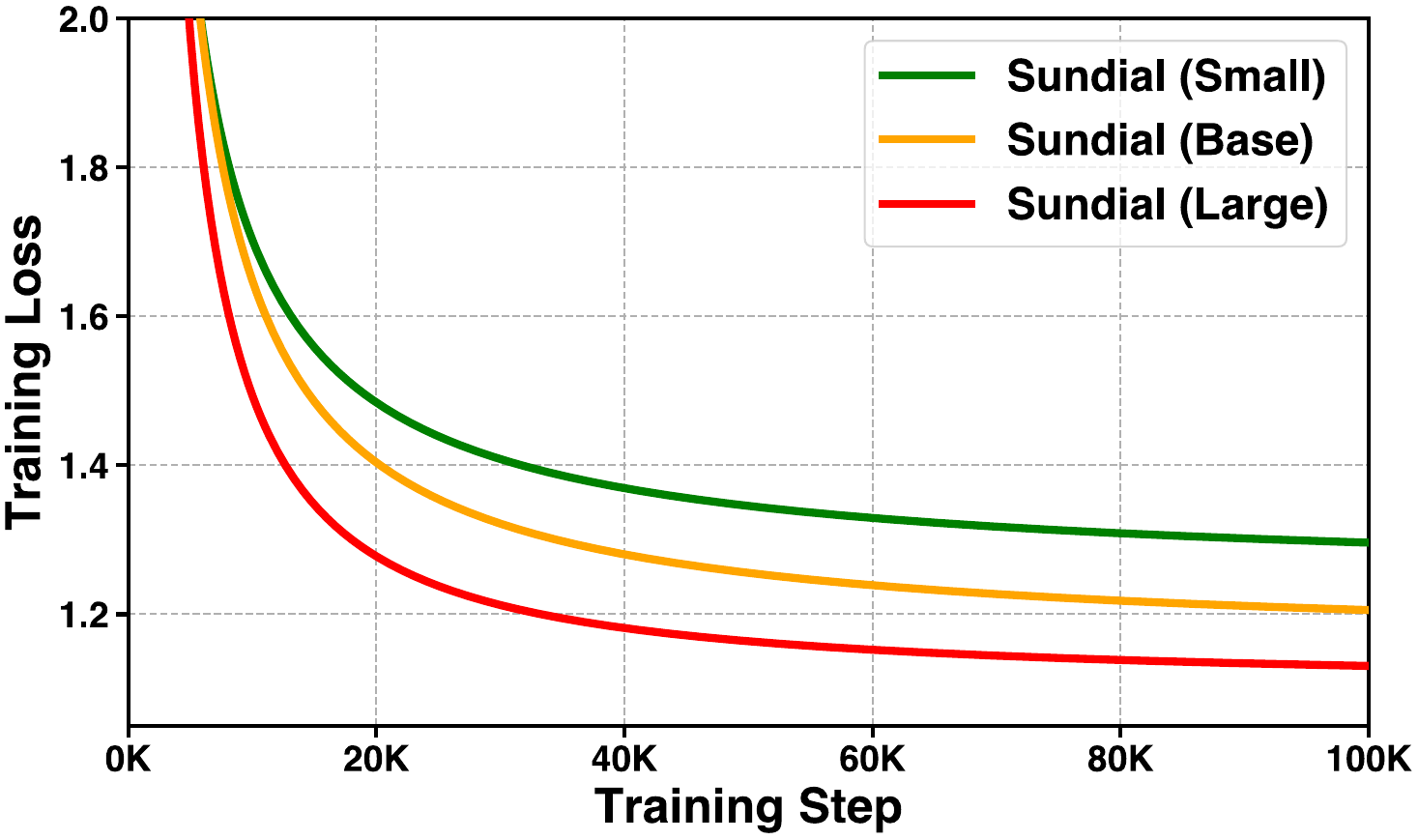}}
    \vspace{-8pt}
	\caption{Training curves on TimeBench of different model sizes.}
	\label{fig:scale_model}
\end{center}
\vspace{-18pt}
\end{figure}

From Table~\ref{tab:zero_shot_datasets}, the larger Sundial model consistently achieves better performance with the scaling of parameters. Beyond downstream performance, we delve into the utilization of model capacity. Figure~\ref{fig:scale_model} shows training curves of different sizes. Compared to Sundial (Small), the large version leads to $15.38\%$ reduction in the converged training loss, exhibiting promising model capacity of generative forecasters.

\subsection{TimeFlow Loss}\label{sec:timeflow}
Based on the flow-matching framework, TimeFlow Loss allows autoregressive models to learn and generate flexible distributions while enhancing representation learning. To validate the effectiveness of this design, we implement two alternatives: (1) an MLP network and MSE Loss and (2) a parameterized training objective based on the denoising diffusion procedure~\cite{li2024autoregressive}. We adopt the same parameterized network and Transformer backbone and pre-train them on TimeBench. Since the converged training loss is not comparable across different objective functions, we compare zero-shot performance in Table~\ref{tab:objective}. Despite allowing for sampling predictions, performance using diffusion-based objective is notably inferior to TimeFlow Loss.

\begin{table}[ht]
  \caption{Zero-shot performance using different training objectives. We use the same model configuration and pre-training scale. Averaged MSE of four prediction lengths are reported here.}
  \label{tab:objective}
  \vspace{-8pt}
  \vskip 0.15in
  \centering
  \renewcommand{\multirowsetup}{\centering}
  \renewcommand\arraystretch{1.2}
  \setlength{\tabcolsep}{2pt}
  \resizebox{\linewidth}{!}{
  \begin{tabular}{c|cccccc|c}
    \toprule
    \scalebox{0.95}{Objective} & \scalebox{0.95}{ETTm1} & \scalebox{0.95}{ETTm2} & \scalebox{0.95}{ETTh1} & \scalebox{0.95}{ETTh2} & \scalebox{0.95}{ECL} & \scalebox{0.95}{Weather} & \scalebox{0.95}{Avg.} \\
    \toprule
    \scalebox{0.95}{\textbf{TimeFlow}} & \scalebox{0.95}{\boldres{0.336}} & \scalebox{0.95}{\boldres{0.258}} & \scalebox{0.95}{\secondres{0.411}}  & \scalebox{0.95}{\boldres{0.333}}  & \scalebox{0.95}{\boldres{0.169}}  & \scalebox{0.95}{\secondres{0.234}} & \scalebox{0.95}{\boldres{0.290}}\\
    \scalebox{0.95}{Diffusion} &  \scalebox{0.95}{0.362}  & \scalebox{0.95}{0.265}  & \scalebox{0.95}{0.444}  & \scalebox{0.95}{0.360}  & \scalebox{0.95}{0.202}  & \scalebox{0.95}{0.252} & \scalebox{0.95}{0.314} \\
    \scalebox{0.95}{MSE} &  \scalebox{0.95}{\secondres{0.360}}  & \scalebox{0.95}{\secondres{0.264}}  & \scalebox{0.95}{\boldres{0.404}}  & \scalebox{0.95}{\secondres{0.341}}  & \scalebox{0.95}{\secondres{0.175}}  & \scalebox{0.95}{\boldres{0.231}}  & \scalebox{0.95}{\secondres{0.296}}\\
    \bottomrule
  \end{tabular}}
  \vspace{-5pt}
\end{table}

In addition to zero-shot performance, we provide showcases for quality evaluations in Appendix~\ref{app:showcase_compare}. Pre-trained models optimized by the specific MSE Loss can only output a single prediction. And the prediction is sometimes over-smooth due to mode collapse (refer to Appendix~\ref{app:mode}). Instead, generative modeling can accommodate significantly different future variations even if their lookback series are similar. We also provide a probablistic metric CRPS to compare different objectives in Table~\ref{tab:crps}, which validate that the predictive distribution modeled by TimeFlow is more coherent and diverse than counterpart training objectives. It benefits downstream tasks by generating multiple plausible predictions, conveys various future possibilities and enhances the reliability of decision-making.

\subsection{Test-Time Calibration}\label{sec:inference}
\begin{figure}[ht]
\begin{center}
\centerline{\includegraphics[width=\columnwidth]{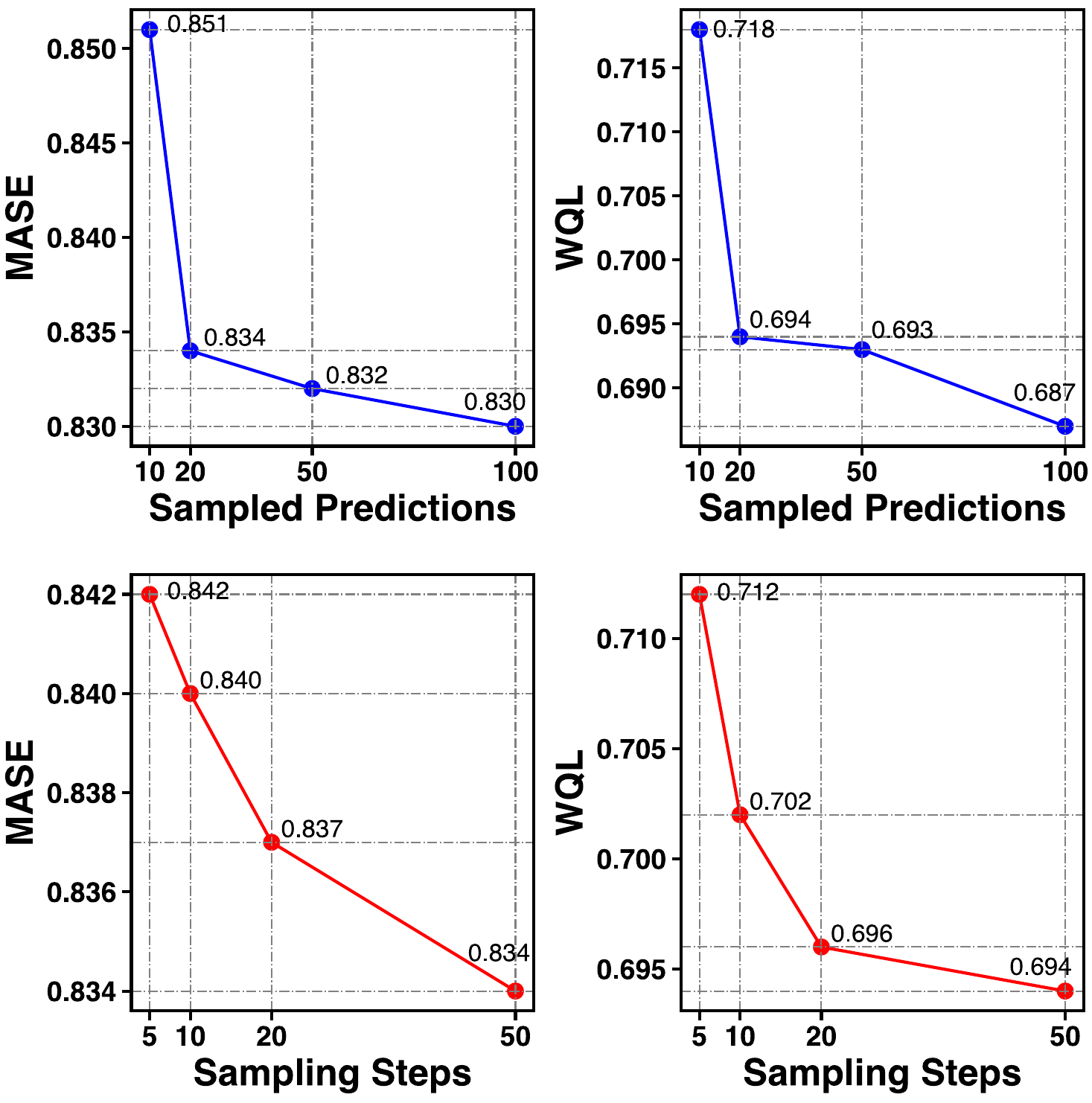}}
    \vspace{-5pt}
	\caption{We show the MASE (left) and WQL (right) on FEV w.r.t. the number of generated raw predictions (top) and the steps to sample a prediction (down). More predictions or more sampling steps generally achieve better probabilistic metrics.}
	\label{fig:inference_tradeoff}
\end{center}
\vspace{-14pt}
\end{figure}

Generative modeling facilitates the flexibility to calibrate the final prediction during inference. Based on the median-based forecasting strategy, i.e., starting from multiple noise of a standard Gaussian and calculating the median of raw predictions, there are two configurations to calibrate final predictions: (1) the number of samples to calculate statistics and (2) sampling steps $K$ used for flow-matching. Figure~\ref{fig:inference_tradeoff} shows the results using different configurations.

The top two figures conform to the central limit theorem. Generating more samples leads to more calibrated estimation of prediction and confidence interval. The bottom two figures indicate that using fine-grained steps during the push-forward process can leads to more precise predictions.

The trade-off between inference time and performance reveals the potential of test-time calibration, which does not require retraining models. The generative capability of Sundial provides flexibility for various use cases requiring different levels of uncertainty. In our experiments, sampling $20$ predictions with each generated by $50$ steps consumes nearly one second on a CPU, which is notably more efficient than tuning deep models or statistical methods. Advanced strategies of sampling and post-processing of raw prediction leave interesting directions for future exploration.

\subsection{Model Adaptation}\label{sec:adapt}

\begin{figure}[ht]
\begin{center}
    \centerline{\includegraphics[width=\columnwidth]{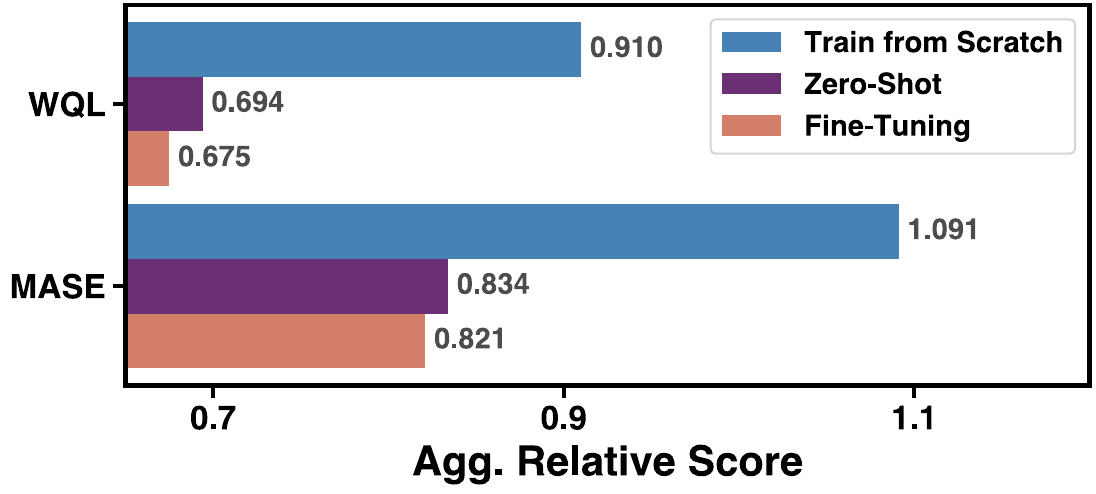}}
    \vspace{-5pt}
	\caption{Performance on the FEV leaderboard, including (1) training Sundial from scratch on all datasets from the FEV leaderboard, (2) zero-shot forecasting using pre-trained Sundial, and (3) fine-tuning once on all datasets from the FEV leaderboard.}
	\label{fig:fine_tune}
\end{center}
\vspace{-25pt}
\end{figure}

Inspired by the prevalence of instruction tuning~\cite{wei2021finetuned} that adapts foundation models on a collection of tasks. We fine-tune pre-trained Sundial (Base) on the FEV leaderboard, including short-term tasks with different prediction lengths. Our model is tuned once on all aggregated datasets. We evaluate the performance on unseen test splits (Figure~\ref{fig:fine_tune}). We observe that the performance can be further improved compared to zero-shot forecasting. Furthermore, training from scratch on aggregated datasets results in inferior performance, implying knowledge transfer in pre-trained models.

\begin{figure*}[ht]
\begin{center}
    \center{\includegraphics[width=\textwidth]{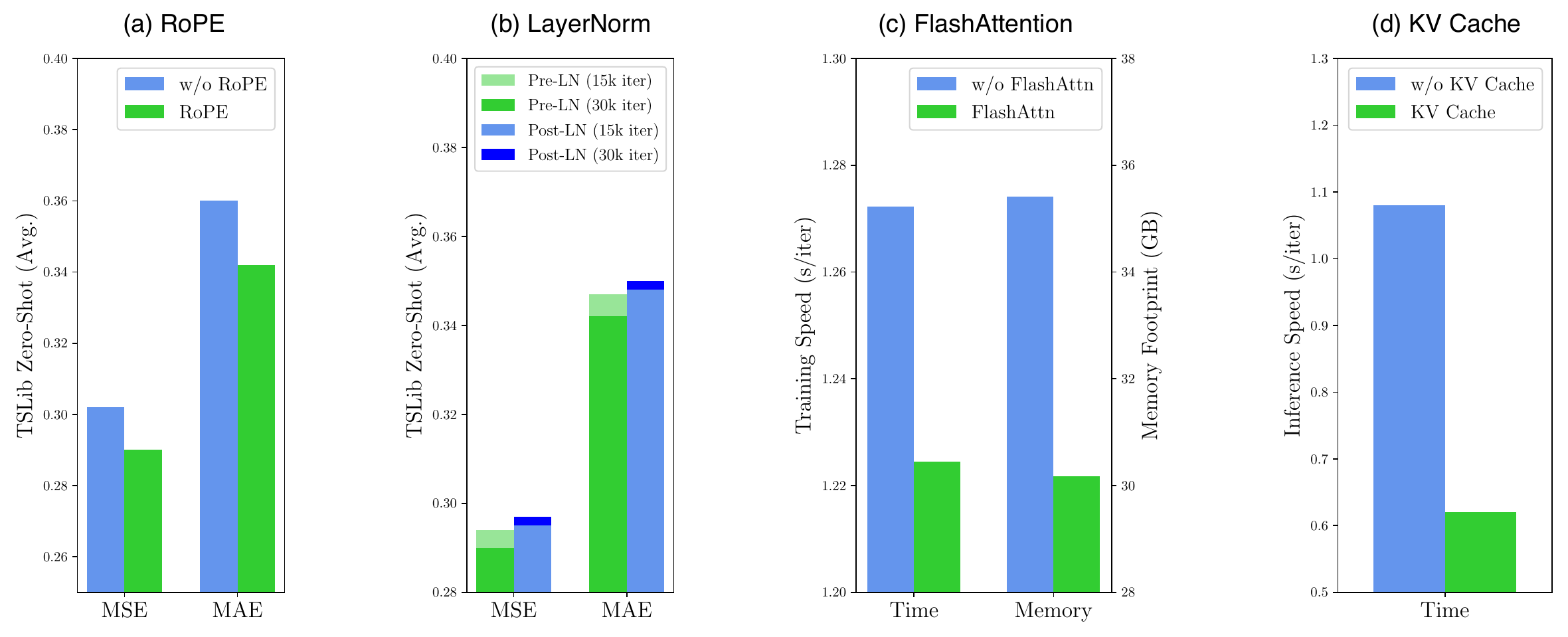}}
    \vspace{-20pt}
	\caption{Ablation studies with respect to architectural enhancements. We report the averaged results of TSLib datasets~\cite{wu2022timesnet} from four prediction lengths $\{96, 192, 336, 720\}$ and all six datasets. The context length is set to $2880$ and the patch length is $16$.}
	\label{fig:ablations}
\end{center}
\vspace{-0pt}
\end{figure*}

\subsection{Ablation Study}\label{sec:ablation}

We conducted several ablation studies that provide insights into the enhancement made to Sundial's architecture. We evaluate the overall zero-shot performance on TSLib, which covers six different datasets and four prediction lengths. 

\paragraph{RoPE} Prior research~\cite{liu2024timer} observed that the introduction of RoPE~\cite{su2024roformer} yields better results in supervised forecasting tasks. As shown in Figure~\ref{fig:ablations} (a), RoPE can also improve zero-shot forecasting, presenting a general enhancement for time series foundation models.

\paragraph{Layer Normalization} Pre-LN~\cite{baevski2018adaptive} is widely adopted in large language models~\cite{touvron2023llama} due to the training stability. As depicted in Figure~\ref{fig:ablations} (b), training with Pre-LN for more iterations yield better performance. In contrast, training with Post-LN, which is the predominant choice in supervised models, may adversely affect downstream results.

\paragraph{FlashAttention and KV Cache} We make it to leverage FlashAttention~\cite{dao2022flashattention} and KV Cache to reduce the computational costs. As shown in Figure~\ref{fig:ablations} (c) and (d), they notably reduce $14.8\%$ memory footprint and $43.6\%$ inference time without affecting performance.

\section{Conclusion}
In this work, we collect and curate TimeBench, a trillion-scale time series dataset for building time series foundation models, which can benefit the research community. Towards time series foundation models, we delve into tokenization and optimization, presenting contributions in two aspects. First, we demonstrate that continuous tokenization, such as patch tokens, can be more effective and efficient for the time series modality, and generative modeling presents a native approach for learning on continuous-valued time series. Second, we propose a novel training objective to accommodate heterogeneous time series distribution. It endows autoregressive models with an inherent capability to sample from non-categorical distribution. Our pre-trained Sundial models make substantial advances on best-recognized forecasting leaderboards. We hope this work can inspire future paradigms for pre-training time series foundation models and enhance their applicability to real-world applications.


\section*{Acknowledgements}
This work was supported by the National Natural Science Foundation of China (U2342217 and 62021002), State Grid Ningxia Electric Power Co. Science and Technology Project (SGNXYX00SCJS2400058), and the BNRist Innovation Fund (BNR2024RC01010), the National Engineering Research Center for Big Data Software.

We extend our gratitude to Xingzhuo Guo for his expertise in flow-matching and meticulous proofreading of the method section. We further thank Jialong Wu, Yuezhou Ma and Yu Zhang for insightful discussions about generative models. Their collective support significantly enhanced this work.

\section*{Impact Statement}
This paper aims to advance the development of time series foundation models. We curated a pre-training dataset from publicly available resources. Our models employ an efficient tokenization and incorporate a generative training objective. The proposed TimeFlow Loss provides insights for training generative foundation models for time series forecasting. We released our pre-trained models that demonstrate notable zero-shot forecasting performance. The generative forecasting paradigm enhances the model reliability for decision-making. Our paper mainly focuses on scientific research and has no obvious negative social impact.

\bibliography{example_paper}
\bibliographystyle{icml2025}

%%%%%%%%%%%%%%%%%%%%%%%%%%%%%%%%%%%%%%%%%%%%%%%%%%%%%%%%%%%%%%%%%%%%%%%%%%%%%%%
%%%%%%%%%%%%%%%%%%%%%%%%%%%%%%%%%%%%%%%%%%%%%%%%%%%%%%%%%%%%%%%%%%%%%%%%%%%%%%%
% APPENDIX
%%%%%%%%%%%%%%%%%%%%%%%%%%%%%%%%%%%%%%%%%%%%%%%%%%%%%%%%%%%%%%%%%%%%%%%%%%%%%%%
%%%%%%%%%%%%%%%%%%%%%%%%%%%%%%%%%%%%%%%%%%%%%%%%%%%%%%%%%%%%%%%%%%%%%%%%%%%%%%%
\newpage
\appendix
\onecolumn

\section{Dataset Statistics}

Large-scale datasets are of paramount importance for pre-training foundation models. Recent research has contributed significant time series datasets~\cite{das2023decoder, liutimer, shi2024time}. While the scaling law of time series foundation models has been explored in the recent work~\cite{shi2024scaling}, the pre-training scale remains relatively limited. Given the heterogeneity of time series compared to other modalities, it raises the question of whether it is feasible to learn from enormous series. To address the question, we curated TimeBench with a trillion time points from various domains.

Unlike other modalities, most time series are unavailable on open websites or repositories. There are also limited domains that encompass typical and predictable time series, leading to slow progress on dataset construction.  Therefore, we conducted tedious preprocessing, including missing values imputation, abnormalities exclusion, and normalization techniques. We conducted statistical analysis, examining time series through the lenses of intrinsic properties, e.g., non-stationarity, forecastability, and seasonality. This approach allows us to characterize the data quality inherent to time series, which affects the training stability of next-token prediction. We also adopt synthetic techniques to improve pattern diversity. Further, we adopt ERA5~\cite{munoz2021era5}, including systematic real-world temporal observations.

The statistical details of TimeBench are summarized in Table~\ref{tab:dataset_summary}. In addition to open-source datasets from research teams on time series foundation models~\cite{woo2024unified, ansari2024chronos, liutimer, liu2024timer}, we collected substantial real-world time series from various domains such as finance, IoT, meteorology, and healthcare~\cite{goldberger2000physiobank}. These resources enable us to construct large-scale time-series corpora exceeding a trillion time points. The corpora include highly credible and predictable data with a wide range of frequencies, lengths, and numbers of variates, providing comprehensive temporal dynamics and variation patterns to facilitate downstream applications. To prevent data leakage, we exclude all datasets evaluated in Section~\ref{sec:zsf} to make sure that Sundial conducts zero-shot forecasting.

\begin{table*}[ht]
    \caption{Key statistics of TimeBench, the pre-training dataset of Sundial.}
    \label{tab:dataset_summary}
    \centering
    \vskip 0.05in
    \renewcommand{\multirowsetup}{\centering}
    \setlength{\tabcolsep}{2pt}
    \resizebox{\linewidth}{!}{
    \begin{tabular}{l|cccccccccccc|c}
        \toprule
        \multirow{2}{*}{\textbf{Source}}
            & Chronos & ECG & Finance &  IoT & LOSTA & Synthetic & ERA5 3h & ERA 12h & ERA5 Daily & \scalebox{1}{ERA5 Weekly} & \scalebox{1}{ERA5 Monthly} & \scalebox{1}{ERA5 Quarterly} & \multirow{2}{*}{\textbf{Total}}\\
            & \citeyearpar{ansari2024chronos} & \citeyearpar{goldberger2000physiobank} & (Ours) & (Ours) & \citeyearpar{woo2024unified} & \citeyearpar{ansari2024chronos} & \citeyearpar{munoz2021era5} & \citeyearpar{munoz2021era5} & \citeyearpar{munoz2021era5} & \citeyearpar{munoz2021era5} & \citeyearpar{munoz2021era5} & \citeyearpar{munoz2021era5} \\
        \midrule
            \textbf{\# Pts.} & 94B & 48B & 10.5B & 5.8B & 230B  & 0.5B & 129B & 32B & 406B & 58B & 13.5B & 4.5B & 1032B\\
            \textbf{\%}  & 9.11 \% & 4.65 \% & 1.02 \% & 0.56 \% & 22.29 \% & 0.05 \% & 12.50 \% & 3.10 \% & 39.35 \% & 5.62 \% & 1.31 \% & 0.44 \% & 100\% \\
        \bottomrule
    \end{tabular}}
\end{table*}

\section{Implementation Details}
All experiments are implemented using PyTorch~\cite{paszke2019pytorch} and executed with $32$ NVIDIA A100 GPUs. We employ the AdamW optimizer~\cite{kingma2014adam} for model optimization. We adopt S3 format~\cite{liutimer} for univariate pre-training. During training, data from different domains is sampled according to a predefined ratio to balance the domain weightings and ensure diversity in the training data. We implement a global shuffle strategy by loading time series into a standard parquet format. We use variable-wise normalization to unify the scope of values.

On the FEV leaderboard~\cite{ansari2024chronos}, which consists of short-term forecasting datasets, we train Sundial models by TimeFlow Loss with the prediction length of $F=16$. For the point forecasting~\cite{wu2022timesnet} and GIFT-Eval~\cite{aksugift}, which consist of forecasting datasets with a prediction length ranging from $6$ to $900$, we train Sundial models by TimeFlow Loss with the prediction length of $F=720$. For the required prediction length less than the model prediction length, we truncate the output generated by Sundial. For the required length more than the prediction horizon, we conduct rolling forecasting. Following Chronos~\cite{ansari2024chronos}, we sample $20$ raw predictions to calculate MASE and WQL on FEV. Being consistent to Moirai~\cite{woo2024unified}, we sample $100$ raw predictions to calculate MASE and CRPS for GIFT-Eval. The sampling step is fixed as $K=50$. Configurations of Sundial in different sizes are provided in Table~\ref{tab:configuration}. We provide a model summary in Table~\ref{tab:ltsm_comp}, which summarizes several aspects of current time series foundation models.

\begin{table*}[ht]
  \caption{Model configurations of the Sundial family.}
  \vspace{-5pt}
  \label{tab:configuration}
  \vskip 0.15in
  \centering
  \renewcommand{\multirowsetup}{\centering}
  \setlength{\tabcolsep}{3pt}
  \renewcommand\arraystretch{1.2}
  \begin{tabular}{c|cccccccc}
    \toprule
    \multirow{2}{*}{Model} & \scalebox{0.9}{Patch Size} & \scalebox{0.9}{Context Length} & \scalebox{0.9}{Prediction Length}  & \scalebox{0.9}{Layers} & \scalebox{0.9}{Dimension} & \scalebox{0.9}{MHA Heads} & \scalebox{0.9}{TimeFlow} & \scalebox{0.9}{Total Parameters} \\
     & $(P)$ & $(T)$ & $(F)$  & $(L)$ & $(D, D_\text{ff})$ & $H$ & $(D_\text{tf}, L_\text{tf})$ & $\#\text{Count}$ \\
    \midrule
    $\textbf{Sundial}_{\textit{Small}}$ & $16$ & $2880$ & $\{16, 720\}$ & 6 & $(512, 2048)$ & $8$ & $(512, 3)$& $32$M \\
    \midrule
    $\textbf{Sundial}_{\textit{Base}}$ &  $16$ & $2880$ & $\{16, 720\}$ & 12 & $(768, 3072)$ & $12$ & $(768, 3)$& $128$M \\
    \midrule
    $\textbf{Sundial}_{\textit{Large}}$ & $16$ & $2880$ & $\{16, 720\}$ & 24 & $(1024, 4096)$ & $16$ & $(1024, 6)$& $444$M \\
    \bottomrule
  \end{tabular}
    \begin{tablenotes}
        \footnotesize
        \item[] $\ast$ 
        $D$ is the embedding dimension of Transformer. $D_{\text{ff}}$ is the hidden dimension of FFN. $D_{\text{tf}}$ is the hidden dimension of the flow-matching network. $L$ is the layer number of Transformer. $L_{\text{tf}}$ is the layer number of the flow-matching network.
  \end{tablenotes}
  \vspace{-5pt}
\end{table*}

\begin{table*}[ht]
  \caption{Comparison of time series foundation models. \emph{Architecture} denotes the Transformer category. \emph{Model Size} presents parameter counts of different model sizes. \emph{Pre-training Scale} measures pre-training datasets in time points. \emph{Token Level} presents the graininess of time series tokens. \emph{Tokenization} denotes what kind of values are embedded from time series. \emph{Context Length} means the input length supported by the model. \emph{Probabilistic} means generating multiple probable predictions, which is the opposite of deterministic forecasters.}
  \vspace{-5pt}
  \label{tab:ltsm_comp}
  \vskip 0.15in
  \centering
  \renewcommand{\multirowsetup}{\centering}
  \setlength{\tabcolsep}{2.5pt}
  \renewcommand\arraystretch{1.2}
  \begin{tabular}{c|ccccccccc}
    \toprule
    \multirow{2}{*}{Method} & \textbf{Sundial}& Time-MoE  & Timer-XL& Moirai & MOMENT & LLMTime & Chronos & Lag-Llama & TimesFM  \\ 
     & \textbf{(Ours)} & \citeyearpar{shi2024time} & \citeyearpar{liu2024timer} &
     \citeyearpar{woo2024unified} & 
     \citeyearpar{goswami2024moment} & \citeyearpar{gruver2024large}  & \citeyearpar{ansari2024chronos} &
     \citeyearpar{rasul2023lag} &  \citeyearpar{das2023decoder}   \\
    \toprule
    Architecture & Decoder & Decoder & Decoder & Encoder & Encoder & Decoder & EncDec & Decoder & Decoder \\
    \midrule
    \multirow{3}{*}{Model Size} & 32M & 113M &84M & 14M & 40M & -  & 46M  & 200M  & 17M  \\
    & 128M & 453M &  & 91M & 125M &   & 200M  &  & 70M   \\
    & 444M & 2.4B &  & 311M & 385M &   & 710M & & 200M  \\
    \midrule
    Pre-training Scale & 1032B & 300B & 260B & 231B & 1.13B & - & 84B & 0.36B & 100B  \\
    \midrule
    Token Level & Patch  & Point & Patch & Patch & Patch & Point & Point & Point & Patch  \\
    \midrule
    Tokenization & \scalebox{0.82}{Continuous}  & \scalebox{0.82}{Continuous} & \scalebox{0.82}{Continuous} & \scalebox{0.82}{Continuous} & \scalebox{0.82}{Continuous} & \scalebox{0.82}{Discrete} & \scalebox{0.82}{Discrete} & \scalebox{0.82}{Continuous} & \scalebox{0.82}{Continuous}  \\
    \midrule
    Context Length & $\le$2880 & $\le$4096 & $\le$2880 & $\le$5000 & = 512 & - & $\le$512 & $\le$1024 & $\le$512 \\
    \midrule
    Probabilistic & True & False & False & True & False & True & True & True & False  \\
    \bottomrule
  \end{tabular}
  \vspace{-5pt}
\end{table*}

\section{Supplementary Results}

\subsection{Discussion of Mode Collapse}\label{app:mode}
Mode collapse is a failure of representation learning, where a model generates a limited variety of outputs, ignoring the diversity in the training data. For time series foundation models, mode collapse stems from the heterogeneity of the time series distribution, e.g., a similar lookback time series goes into divergent trending. In other words, the semantics of time series patterns are highly unstable. It sometimes leads to over-smooth predictions from models optimized by MSE becuase the results are global-optimal for this loss (See showcases on the right of Figure~\ref{fig:showcases_compare1}-\ref{fig:showcases_compare2}). Such a training objective pre-defines a unimodal predictive distribution of data, which struggles to accommodate large-scale datasets like TimeBench.

Our work addresses this phenomenon through generative modeling. Generative forecasters learn flexible distributions without relying on probabilistic priors. We evaluate the distributional metric Continuous Ranked Probability Score (CRPS) to assess the quality of generated predictions across different training objectives. The results indicate that the predictive distribution modeled by TimeFlow is more coherent and diverse compared to alternative training objectives, particularly on the highly diverse GIFT-Eval~\cite{aksugift}. It validates the effectiveness of TimeFlow in mitigating mode collapse.

\begin{table*}[ht]
  \vspace{-15pt}
  \caption{Zero-shot probabilistic forecasting performance using different training objectives. Averaged CRPS is reported here.}
  \vspace{-5pt}
  \label{tab:crps}
  \vskip 0.15in
  \centering
  \begin{small}
  \renewcommand{\multirowsetup}{\centering}
  \setlength{\tabcolsep}{14.5pt}
  \renewcommand\arraystretch{1.2}
  \begin{tabular}{c|ccccccc}
    \toprule
    \multirow{1}{*}{Objective} & ETTh1 & ETTh2 & ETTm1 & ETTm2 & ECL & Weather & GIFT-Eval   \\ 
    \toprule
    \multirow{1}{*}{TimeFlow Loss}  & \textbf{0.0059} & \textbf{0.0037} & \textbf{0.0057} & \textbf{0.0029} & 0.0082 & \textbf{0.0021} & \textbf{0.5050} \\
    \midrule
    \multirow{1}{*}{Diffusion Loss} & 0.0082 & 0.0053 & 0.0070 & 0.0039 & 0.0095 & 0.0032 & 0.5340 \\
    \midrule
    \multirow{1}{*}{MSE Loss}       & 0.0063 & 0.0040 & 0.0058 & 0.0032 & \textbf{0.0080} & 0.0023 & 0.6420 \\
    \bottomrule
  \end{tabular}
  \end{small}
  \vspace{-5pt}
\end{table*}

\subsection{Scaling Behavior Using More Data}
We compare Sundial with other time series foundation models that are pre-trained with smaller datasets: Chronos~\cite{ansari2024chronos} is pre-trained on 94 billion time points, and Moirai is pre-trained on 230 billion time points. As their pre-training datasets are part of the subset of TimeBench, we also conduct pre-training on Sundial using these subsets. As shown in Table~\ref{tab:data_scale}, these results highlight the scaling behavior of Sundial using larger datasets. Additionally, Sundial still achieves better zero-shot forecasting than its counterpart models with the same pre-training dataset.

\begin{table*}[ht]
  \vspace{-5pt}
  \caption{Zero-shot forecasting performance of models trained on different scales of datasets (measured in time points, pts, and 1B means a billion). We report the averaged results from four prediction lengths $\{96, 192, 336, 720\}$ on Time-Series-Library~\cite{wu2022timesnet}.}
  \vskip 0.1in
  \label{tab:data_scale}
  \footnotesize
  \begin{small}
  \linespread{2}
  \renewcommand{\multirowsetup}{\centering}
  \setlength{\tabcolsep}{10.5pt}
  \renewcommand\arraystretch{1.4}
  \begin{tabular}{c|cc|cc|cc|cc|cc}
    \toprule
    Model (pts.) & \multicolumn{2}{c|}{Chronos (94B)} & \multicolumn{2}{c|}{Moirai (230B)} & \multicolumn{2}{c|}{Sundial (94B)} & \multicolumn{2}{c|}{Sundial (230B)} & \multicolumn{2}{c}{Sundial (1032B)} \\
    \cmidrule(lr){1-1} \cmidrule(lr){2-3} \cmidrule(lr){4-5} \cmidrule(lr){6-7} \cmidrule(lr){8-9} \cmidrule(lr){10-11}
    Dataset & MSE & MAE & MSE & MAE &MSE & MAE &MSE & MAE &MSE & MAE  \\
    \toprule
    ETTh1  & 0.591 & 0.468 & 0.417 & \textbf{0.419}  & 0.402 & 0.429 & 0.403 & \textbf{0.419}	& 0.411 & 0.434\\
    \midrule
    ETTh2  & 0.405 & 0.410 & 0.362 & \textbf{0.382}  	& 0.377 & 0.414 & 0.364 & 0.398	& \textbf{0.333} & 0.387\\
    \midrule
    ETTm1  & 0.645 & 0.500 & 0.406 & 0.385  & 0.367 & 0.402 & 0.352 & 0.385	& \textbf{0.336} & \textbf{0.377} \\
    \midrule
    ETTm2  & 0.310 & 0.350 & 0.311 & 0.337 	& 0.280 & 0.341 & 0.273 & 0.334	& \textbf{0.258} & \textbf{0.320}\\
    \midrule    
    ECL   & 0.214 & 0.278 & 0.187 & 0.274   	& 0.172 & 0.269 & 0.171 & 0.267	& \textbf{0.169} & \textbf{0.265}\\
    \midrule
    Weather  & 0.292 & 0.315 & 0.287 & 0.281	& 0.254 & 0.301 & 0.252 & 0.297	& \textbf{0.234} & \textbf{0.270}\\
    \bottomrule
  \end{tabular}
  \end{small}
\end{table*}

\subsection{Performance with Varying Lookback Lengths}
\begin{figure*}[ht]
\begin{center}
    \center{\includegraphics[width=\textwidth]{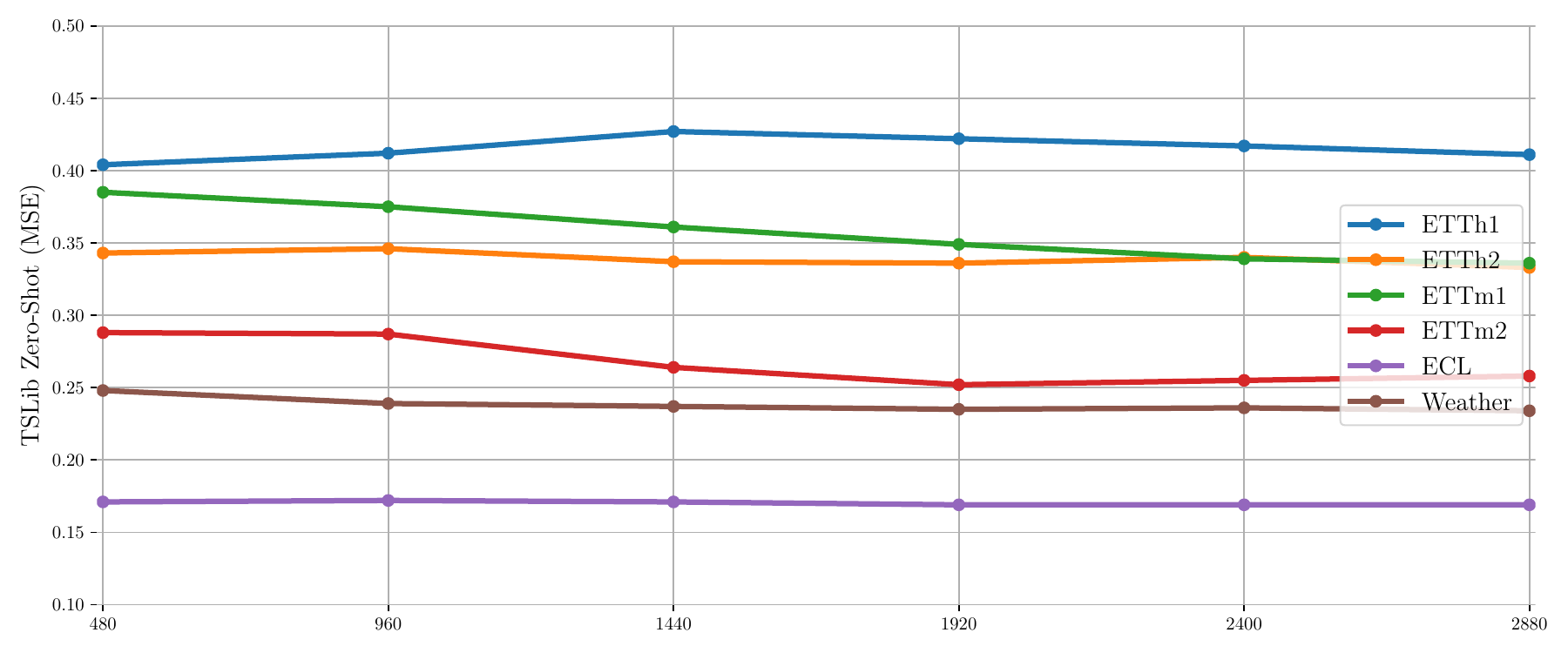}}
    \vspace{-10pt}
	\caption{Zero-shot forecasting performance using different lookback lengths in $\{480, 960, 1440, 1920, 2400, 2880\}$. We report the averaged results from four prediction lengths $\{96, 192, 336, 720\}$ on Time-Series-Library~\cite{wu2022timesnet}.}
	\label{fig:lookback}
\end{center}
\end{figure*}

Time series foundation models operate independently of training, functioning similarly to statistical methods. Given specific forecasting tasks, one of the most important hyperparameters is the lookback length. Unlike fixed-context models, Sundial offers flexibility for practitioners, allowing the context length to be dynamically adjusted during inference. In Figure~\ref{fig:lookback}, we present the performance of Sundial utilizing various lookback lengths. Based on our observations, we contend that performance is largely dependent on the forecasting task itself. Specifically, the size of the lookback window can be tuned to meet the forecasting horizon and data periodicity. Time series foundation models provide a training-free approach for rapid adjustments; still, they should enhance their fundamental long-context capabilities to handle high-frequency data.

\subsection{Zero-Shot Results of Point Forecasting}
Table~\ref{tab:zero_shot_datasets_full} provides full zero-shot results on Time-Series-Library forecasting benchmark~\cite{wu2022timesnet}, including prediction horizons in $\{96, 192, 336, 720\}$. We build Sundial with different model sizes with configurations in Table~\ref{tab:configuration}. The context length is fixed as $2880$. We truncate the model's predictions for tasks requiring a prediction length less than $F=720$.

We compare the most advanced time series foundation models based on their official checkpoints, including Time-MoE~\cite{shi2024time},  Timer~\cite{liu2024timer, liutimer}, Moirai~\cite{woo2024unified}, TimesFM~\cite{das2023decoder}, and Chronos~\cite{ansari2024chronos}. We conduct zero-shot evaluations on datasets that are not included during the pre-training of the corresponding models. For each of the evaluated model, we use their maximum input length during inference. Metrics (MSE/MAE) are calculated from all predicted windows in the test split of each dataset following~\citet{liu2024timer}.

\subsection{Zero-Shot Results on GIFT-Eval and FEV Leaderboard}
We evaluate our models on GIFT-Eval, a benchmark designed to comprehensively assess forecasting performance across diverse time series. GIFT-Eval includes $23$ datasets covering $144,000$ time series and $177$ million data points, which constitute a total of $97$ forecasting configurations. We use the official evaluation suite established by the research team of Salesforce and report aggregated results in Table~\ref{tab:gift_eval}. We evaluate the performance and inference time on the FEV leaderboard, which was originally proposed by~\citet{ansari2024chronos} and established by AutoGluon, which comprises $27$ datasets for zero-shot evaluation. We report aggregated metrics in Figure \ref{fig:fev} and assess the inference time in Figure \ref{fig:inference_time}. We released the detailed results by submitting Sundial to their open benchmark\footnote{\href{https://huggingface.co/spaces/Salesforce/GIFT-Eval}{https://huggingface.co/spaces/Salesforce/GIFT-Eval}.}.

\section{Showcases}\label{app:showcase}

\subsection{Showcases of Sundial}
\begin{figure*}[ht]
\begin{center}
    \center{\includegraphics[width=\textwidth]{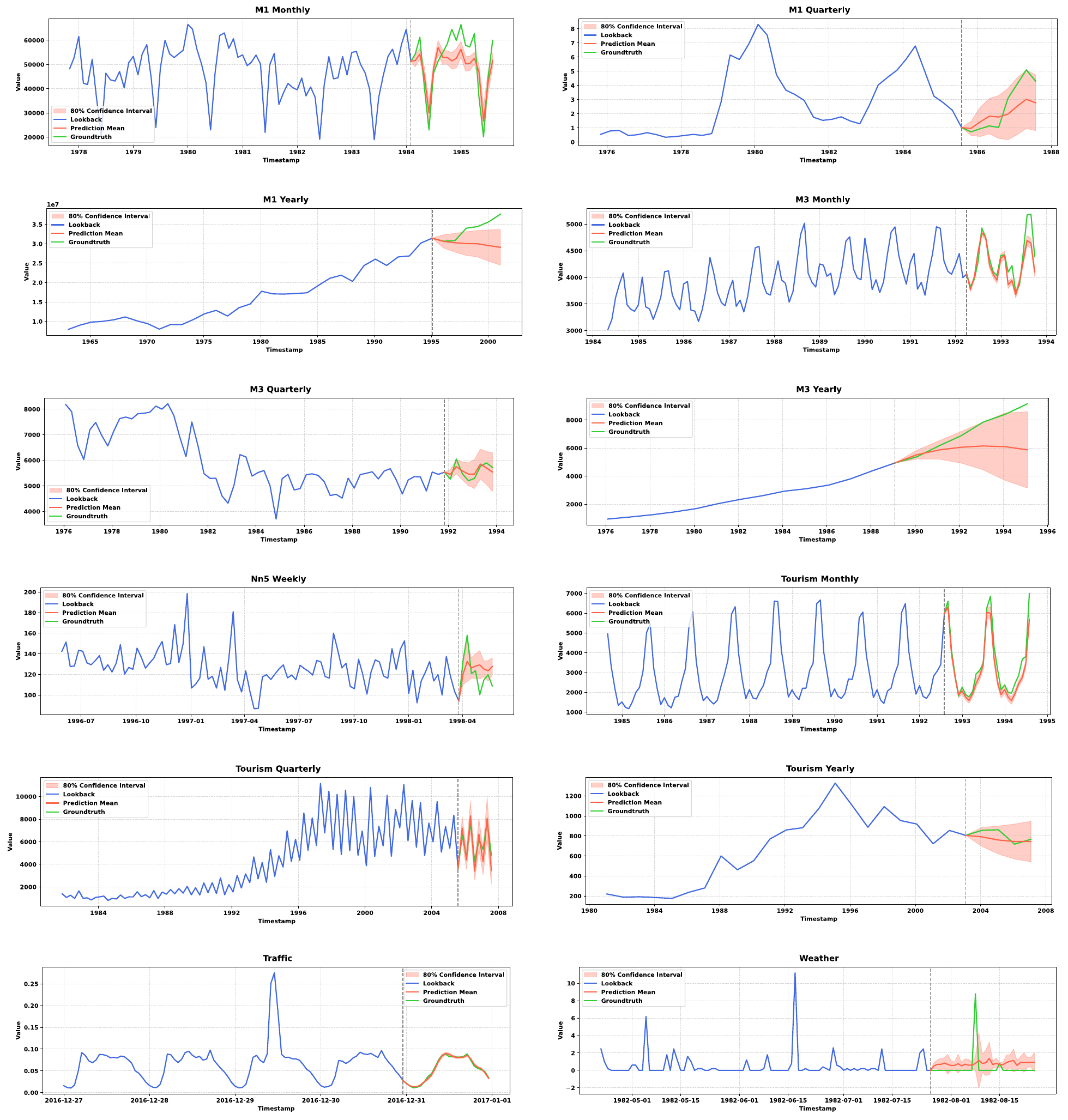}}
    \vspace{-2pt}
	\caption{Showcases of zero-shot predictions from Sundial (Base) on the FEV leaderboard~\cite{ansari2024chronos}.}
	\label{fig:showcases_fev1}
\end{center}
\end{figure*}

\begin{figure*}[ht]
\begin{center}
    \center{\includegraphics[width=\textwidth]{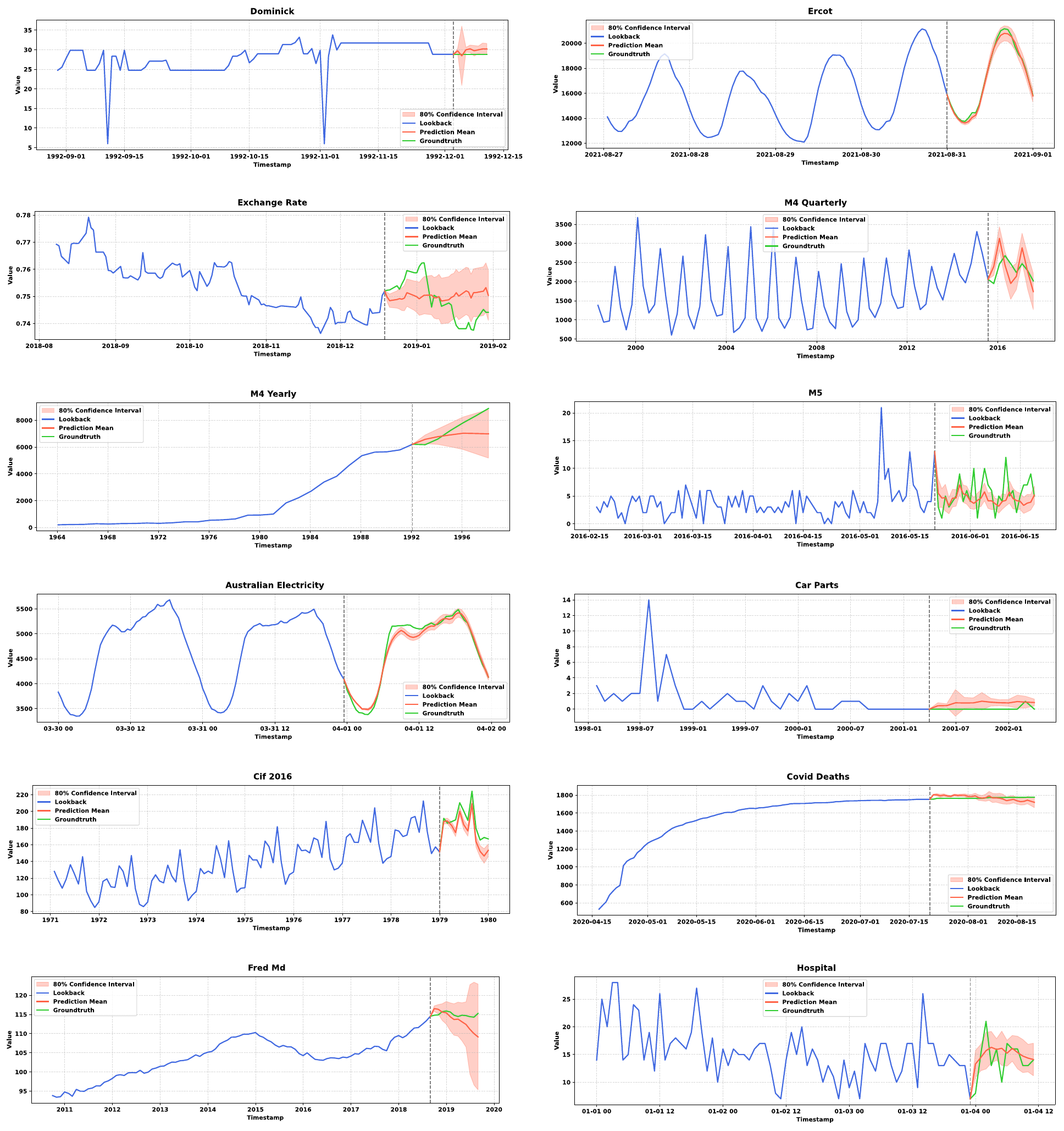}}
    \vspace{-2pt}
	\caption{Showcases of zero-shot predictions from Sundial (Base) on the FEV leaderboard~\cite{ansari2024chronos}.}
	\label{fig:showcases_fev2}
\end{center}
\end{figure*}

\begin{figure*}[ht]
\begin{center}
    \center{\includegraphics[width=\textwidth]{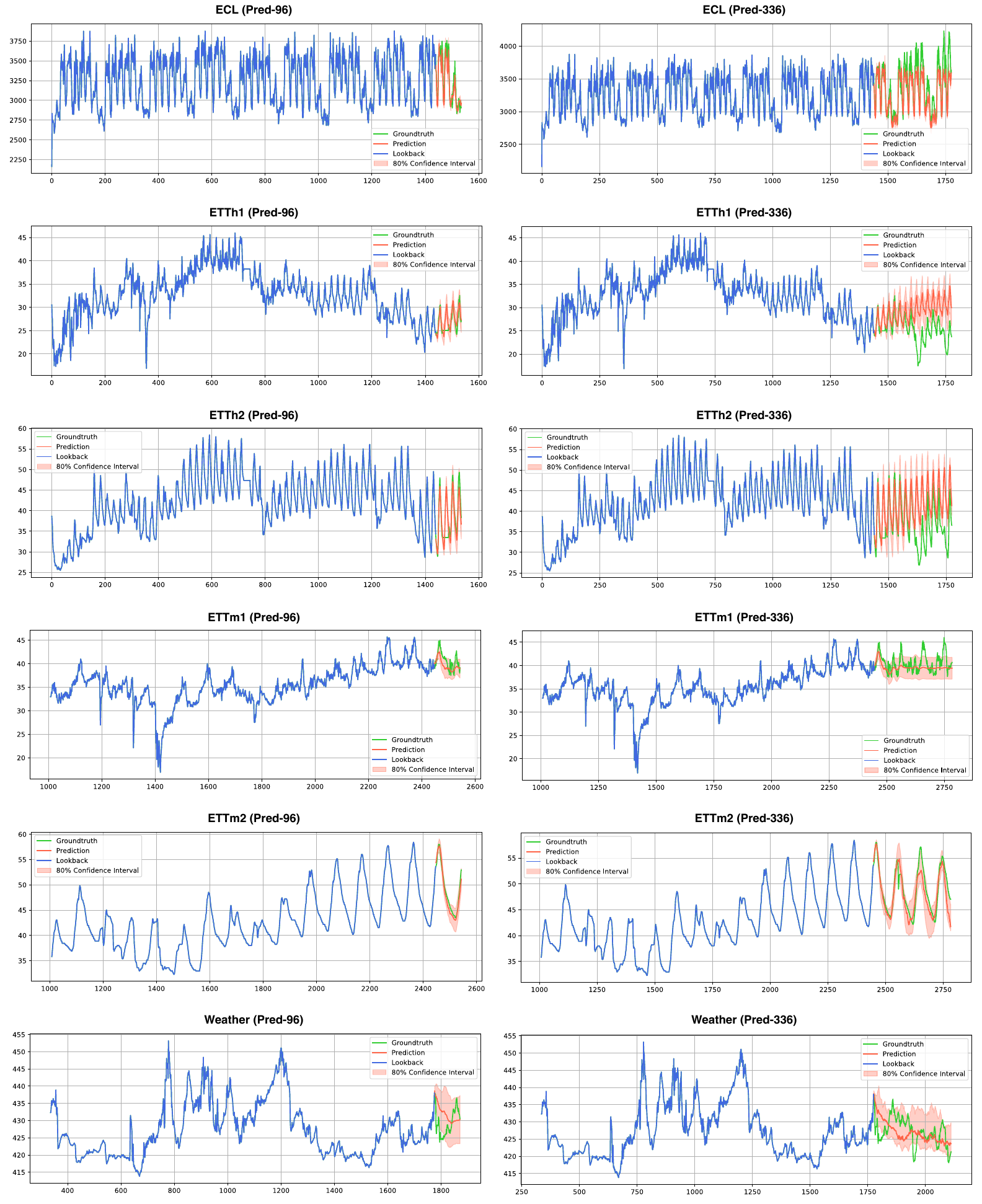}}
    \vspace{-2pt}
	\caption{Showcases of zero-shot predictions from Sundial (Base) on long-term forecasting datasets~\cite{wu2022timesnet}.}
	\label{fig:showcases_tslib}
\end{center}
\end{figure*}

Figure~\ref{fig:showcases_fev1}-\ref{fig:showcases_tslib} present zero-shot forecasting showcases on all the datasets from FEV~\cite{ansari2024chronos} and TSLib~\cite{wu2022timesnet}. By generating $20$ predictions with different initial noise, we estimate the median and $80\%$ prediction interval.

\begin{table}[htbp]
  \vspace{-8pt}
  \caption{Zero-shot forecasting results of time series foundation models on long-term forecasting datasets~\cite{wu2022timesnet}. A lower MSE or MAE indicates a better prediction. Averaged results of four prediction lengths are reported here. $1^{\text{st}}$ Count represents the number of wins achieved by a model under all prediction lengths and datasets. Results of baseline models are officially reported by \citet{shi2024time}. Datasets for pre-training are not evaluated on corresponding models, which are denoted by the dash ($-$).}
  \vspace{-3pt}
  \renewcommand{\arraystretch}{0.89} 
  \centering
  \begin{threeparttable}
  \begin{small}
  \renewcommand{\multirowsetup}{\centering}
  \setlength{\tabcolsep}{1.2pt}
  \label{tab:zero_shot_datasets_full}
  \begin{tabular}{c|c|cc|cc|cc|cc|cc|cc|cc|cc|cc|cc|cc|cc}
    \toprule
    \multicolumn{2}{c}{\multirow{2}{*}{Models}} & 
    \multicolumn{2}{c}{\rotatebox{0}{\scalebox{0.8}{$\textbf{Sundial}_{\textit{Small}}$}}} &
   \multicolumn{2}{c}{\rotatebox{0}{\scalebox{0.8}{$\textbf{Sundial}_{\textit{Base}}$}}} &
   \multicolumn{2}{c}{\rotatebox{0}{\scalebox{0.8}{$\textbf{Sundial}_{\textit{Large}}$}}} &

    \multicolumn{2}{c}{\rotatebox{0}{\scalebox{0.68}{$\textbf{Time-MoE}_{\textit{Base}}$}}} &
    \multicolumn{2}{c}{\rotatebox{0}{\scalebox{0.68}{$\textbf{Time-MoE}_{\textit{Large}}$}}} &
    \multicolumn{2}{c}{\rotatebox{0}{\scalebox{0.68}{$\textbf{Time-MoE}_{\textit{Ultra}}$}}}  &
    \multicolumn{2}{c}{\rotatebox{0}{\scalebox{0.8}{$\textbf{Timer-XL}$}}} &
    \multicolumn{2}{c}{\rotatebox{0}{\scalebox{0.8}{{$\textbf{Moirai}_{\textit{Base}}$}}}} &
    \multicolumn{2}{c}{\rotatebox{0}{\scalebox{0.8}{$\textbf{Moirai}_{\textit{Large}}$}}}&
    \multicolumn{2}{c}{\rotatebox{0}{\scalebox{0.8}{$\textbf{Chronos}_{\textit{Base}}$}}} &
    \multicolumn{2}{c}{\rotatebox{0}{\scalebox{0.8}{$\textbf{Chronos}_{\textit{Large}}$}}} &
    \multicolumn{2}{c}{\rotatebox{0}{\scalebox{0.8}{$\textbf{TimesFM}$}}}\\

    \multicolumn{2}{c}{} &
    \multicolumn{2}{c}{\scalebox{0.8}{(Ours)}} & 
    \multicolumn{2}{c}{\scalebox{0.8}{(Ours)}} &
    \multicolumn{2}{c}{\scalebox{0.8}{(Ours)}} &
    \multicolumn{2}{c}{\scalebox{0.8}{\citeyearpar{shi2024time}}} & 
    \multicolumn{2}{c}{\scalebox{0.8}{\citeyearpar{shi2024time}}} & 
    \multicolumn{2}{c}{\scalebox{0.8}{\citeyearpar{shi2024time}}}  & 
    \multicolumn{2}{c}{\scalebox{0.8}{\citeyearpar{liu2024timer}}} & 
    \multicolumn{2}{c}{\scalebox{0.8}{\citeyearpar{woo2024unified}}} & 
    \multicolumn{2}{c}{\scalebox{0.8}{\citeyearpar{woo2024unified}}}& 
    \multicolumn{2}{c}{\scalebox{0.8}{\citeyearpar{ansari2024chronos}}} &
    \multicolumn{2}{c}{\scalebox{0.8}{\citeyearpar{ansari2024chronos}}} &
    \multicolumn{2}{c}{\scalebox{0.8}{\citeyearpar{das2023decoder}}} \\

    \cmidrule(lr){3-4} \cmidrule(lr){5-6}\cmidrule(lr){7-8} \cmidrule(lr){9-10}\cmidrule(lr){11-12}\cmidrule(lr){13-14} \cmidrule(lr){15-16} \cmidrule(lr){17-18} \cmidrule(lr){19-20} \cmidrule(lr){21-22} \cmidrule(lr){23-24} \cmidrule(lr){25-26}

    \multicolumn{2}{c}{Metric}  & \scalebox{0.78}{MSE} & \scalebox{0.78}{MAE}  & \scalebox{0.78}{MSE} & \scalebox{0.78}{MAE}  & \scalebox{0.78}{MSE} & \scalebox{0.78}{MAE}  & \scalebox{0.78}{MSE} & \scalebox{0.78}{MAE}  & \scalebox{0.78}{MSE} & \scalebox{0.78}{MAE}  & \scalebox{0.78}{MSE} & \scalebox{0.78}{MAE}  & \scalebox{0.78}{MSE} & \scalebox{0.78}{MAE} & \scalebox{0.78}{MSE} & \scalebox{0.78}{MAE} & \scalebox{0.78}{MSE} & \scalebox{0.78}{MAE} & \scalebox{0.78}{MSE} & \scalebox{0.78}{MAE} & \scalebox{0.78}{MSE} & \scalebox{0.78}{MAE} & \scalebox{0.78}{MSE} & \scalebox{0.78}{MAE} \\
    \toprule

    \multirow{5}{*}{\rotatebox{90}{\scalebox{0.95}{ETTm1}}}
    & \scalebox{0.78}{96} &\scalebox{0.78}{0.292} &\scalebox{0.78}{0.342} &\secondres{\scalebox{0.78}{0.280}} &\secondres{\scalebox{0.78}{0.334}} &\boldres{\scalebox{0.78}{0.273}} &\boldres{\scalebox{0.78}{0.329}} & \scalebox{0.78}{0.338} & \scalebox{0.78}{0.368} & \scalebox{0.78}{0.309} & \scalebox{0.78}{0.357} & \scalebox{0.78}{0.281} & \scalebox{0.78}{0.341}& {\scalebox{0.78}{0.317}} & \scalebox{0.78}{0.356}  & {\scalebox{0.78}{0.363}} &\scalebox{0.78}{0.356} &{\scalebox{0.78}{0.380}} &{\scalebox{0.78}{0.361}} & \scalebox{0.78}{0.454} &\scalebox{0.78}{0.408} &\scalebox{0.78}{0.457} &\scalebox{0.78}{0.403} &\scalebox{0.78}{0.361} &\scalebox{0.78}{0.370} \\ 

    & \scalebox{0.78}{192} &\scalebox{0.78}{0.337} &\scalebox{0.78}{0.376}&\scalebox{0.78}{0.321} &\scalebox{0.78}{0.366}&\secondres{\scalebox{0.78}{0.312}} &\boldres{\scalebox{0.78}{0.357}} & \scalebox{0.78}{0.353} & \scalebox{0.78}{0.388} & \scalebox{0.78}{0.346} & \scalebox{0.78}{0.381} & \boldres{\scalebox{0.78}{0.305}} & \secondres{\scalebox{0.78}{0.358}} & \scalebox{0.78}{0.358} & \scalebox{0.78}{0.381}& {\scalebox{0.78}{0.388}} &{\scalebox{0.78}{0.375}} &{\scalebox{0.78}{0.412}} &{\scalebox{0.78}{0.383}} &\scalebox{0.78}{0.567} &\scalebox{0.78}{0.477} &\scalebox{0.78}{0.530} &\scalebox{0.78}{0.450} &\scalebox{0.78}{0.414} &\scalebox{0.78}{0.405} \\ 
    
    & \scalebox{0.78}{336} 
    &\scalebox{0.78}{0.370} 
    &\scalebox{0.78}{0.401}
    &\secondres{\scalebox{0.78}{0.350}} 
    &\secondres{\scalebox{0.78}{0.389}}
    &\boldres{\scalebox{0.78}{0.343}} 
    &\boldres{\scalebox{0.78}{0.378}}
    & \scalebox{0.78}{0.381} 
    & \scalebox{0.78}{0.413} 
    & \scalebox{0.78}{0.373} 
    & \scalebox{0.78}{0.408} 
    & \scalebox{0.78}{0.369}  
    &\scalebox{0.78}{0.395}
    & \scalebox{0.78}{0.386} 
    & \scalebox{0.78}{0.401} 
    &\scalebox{0.78}{0.416} 
    &\scalebox{0.78}{0.392}
    &{\scalebox{0.78}{0.436}} 
    &{\scalebox{0.78}{0.400}}  
    &\scalebox{0.78}{0.662}
    &\scalebox{0.78}{0.525} 
    &\scalebox{0.78}{0.577} 
    &\scalebox{0.78}{0.481} 
    &\scalebox{0.78}{0.445} 
    &\scalebox{0.78}{0.429} \\
    
    & \scalebox{0.78}{720} 
    &\scalebox{0.78}{0.418} 
    &\scalebox{0.78}{0.433}
    &\boldres{\scalebox{0.78}{0.394}} 
    &\secondres{\scalebox{0.78}{0.418}}
    &\secondres{\scalebox{0.78}{0.397}} 
    &\boldres{\scalebox{0.78}{0.413}}

    &\scalebox{0.78}{0.504} 
    &\scalebox{0.78}{0.493} 
    &\scalebox{0.78}{0.475} 
    & \scalebox{0.78}{0.477} 
    & \scalebox{0.78}{0.469} 
    & \scalebox{0.78}{0.472} 
    &\scalebox{0.78}{0.430}
    &\scalebox{0.78}{0.431} 
    &\scalebox{0.78}{0.460}
    &\scalebox{0.78}{0.418}
    &\scalebox{0.78}{0.462} 
    &\scalebox{0.78}{0.420}
    &\scalebox{0.78}{0.900} 
    &\scalebox{0.78}{0.591} 
    &\scalebox{0.78}{0.660} 
    &\scalebox{0.78}{0.526} 
    &\scalebox{0.78}{0.512} 
    &\scalebox{0.78}{0.471} \\

    \cmidrule(lr){2-26}
    & \scalebox{0.78}{Avg} 
    &\scalebox{0.78}{0.354} 
    &\scalebox{0.78}{0.388}
    &\secondres{\scalebox{0.78}{0.336}} 
    &\secondres{\scalebox{0.78}{0.377}}
    &\boldres{\scalebox{0.78}{0.331}} 
    &\boldres{\scalebox{0.78}{0.369}}

    & \scalebox{0.78}{0.394} 
    & \scalebox{0.78}{0.415} 
    & \scalebox{0.78}{0.376} 
    & \scalebox{0.78}{0.405} 
    &\scalebox{0.78}{0.356}
    &\scalebox{0.78}{0.391}
    &\scalebox{0.78}{0.373}
    & \scalebox{0.78}{0.392} 
    & \scalebox{0.78}{0.406} 
    &\scalebox{0.78}{0.385}
    &{\scalebox{0.78}{0.422}} 
    &{\scalebox{0.78}{0.391}}
    &\scalebox{0.78}{0.645} 
    &\scalebox{0.78}{0.500} 
    &\scalebox{0.78}{0.555} 
    &\scalebox{0.78}{0.465} 
    &\scalebox{0.78}{0.433} 
    &\scalebox{0.78}{0.418} \\

    \midrule
    
    \multirow{5}{*}{\rotatebox{90}{\scalebox{0.95}{ETTm2}}}
    &
    \scalebox{0.78}{96} 
    &\scalebox{0.78}{0.178} 
    &\scalebox{0.78}{0.260}
    &\boldres{\scalebox{0.78}{0.170}} 
    &\secondres{\scalebox{0.78}{0.256}}
    &\secondres{\scalebox{0.78}{0.172}} 
    &\boldres{\scalebox{0.78}{0.255}}
    & \scalebox{0.78}{0.201} 
    & \scalebox{0.78}{0.291} 
    &{\scalebox{0.78}{0.197}} 
    & \scalebox{0.78}{0.286} 
    & \scalebox{0.78}{0.198} 
    & \scalebox{0.78}{0.288} 
    &{\scalebox{0.78}{0.189}} 
    & \scalebox{0.78}{0.277} 
    & {\scalebox{0.78}{0.205}} 
    &\scalebox{0.78}{0.273} 
    &\scalebox{0.78}{0.211} 
    &\scalebox{0.78}{0.274} 
    &\scalebox{0.78}{0.199} 
    &\scalebox{0.78}{0.274} 
    &\scalebox{0.78}{0.197}
    &{\scalebox{0.78}{0.271}} 
    &\scalebox{0.78}{0.202}
    &\scalebox{0.78}{0.270} \\ 
    
    &\scalebox{0.78}{192}
    &\scalebox{0.78}{0.235} 
    &\scalebox{0.78}{0.304} 
    &\secondres{\scalebox{0.78}{0.229}} 
    &\secondres{\scalebox{0.78}{0.300}}
    &\boldres{\scalebox{0.78}{0.227}}
    &\boldres{\scalebox{0.78}{0.296}}
    & \scalebox{0.78}{0.258} 
    & \scalebox{0.78}{0.334} 
    & \scalebox{0.78}{0.250} 
    & \scalebox{0.78}{0.322} 
    &\scalebox{0.78}{0.235}
    &\scalebox{0.78}{0.312}
    &\scalebox{0.78}{0.241}
    &\scalebox{0.78}{0.315}
    &\scalebox{0.78}{0.275}
    &\scalebox{0.78}{0.316}
    &\scalebox{0.78}{0.281} 
    &\scalebox{0.78}{0.318} 
    & \scalebox{0.78}{0.261} 
    &\scalebox{0.78}{0.322} 
    &\scalebox{0.78}{0.254} 
    &\scalebox{0.78}{0.314}
    &\scalebox{0.78}{0.289} 
    &\scalebox{0.78}{0.321}\\
    
    & \scalebox{0.78}{336} 
    &\scalebox{0.78}{0.287} 
    &\scalebox{0.78}{0.342}
    &\secondres{\scalebox{0.78}{0.281}} 
    &\secondres{\scalebox{0.78}{0.337}}
    &\boldres{\scalebox{0.78}{0.275}} 
    &\boldres{\scalebox{0.78}{0.331}}

    & \scalebox{0.78}{0.324} 
    &\scalebox{0.78}{0.373} 
    & \scalebox{0.78}{0.337} 
    & \scalebox{0.78}{0.375}  
    & \scalebox{0.78}{0.293}
    & \scalebox{0.78}{0.348}
    & \scalebox{0.78}{0.286}
    & \scalebox{0.78}{0.348}
    & \scalebox{0.78}{0.329}
    &\scalebox{0.78}{0.350}
    &\scalebox{0.78}{0.341} 
    &\scalebox{0.78}{0.355} 
    &\scalebox{0.78}{0.326} 
    &\scalebox{0.78}{0.366} 
    &\scalebox{0.78}{0.313} 
    &\scalebox{0.78}{0.353}
    &\scalebox{0.78}{0.360} 
    &\scalebox{0.78}{0.366} \\ 
    
    & \scalebox{0.78}{720} 
    &\scalebox{0.78}{0.360} 
    &\scalebox{0.78}{0.390}
    &\secondres{\scalebox{0.78}{0.351}} 
    &\secondres{\scalebox{0.78}{0.387}}
    &\boldres{\scalebox{0.78}{0.343}} 
    &\boldres{\scalebox{0.78}{0.378}}
    & \scalebox{0.78}{0.488} 
    & \scalebox{0.78}{0.464} 
    & \scalebox{0.78}{0.480} 
    & \scalebox{0.78}{0.461} 
    & \scalebox{0.78}{0.427} 
    & \scalebox{0.78}{0.428} 
    & \scalebox{0.78}{0.375}
    & \scalebox{0.78}{0.402}
    & \scalebox{0.78}{0.437}
    &\scalebox{0.78}{0.411}
    &\scalebox{0.78}{0.485} 
    &\scalebox{0.78}{0.428} 
    &\scalebox{0.78}{0.455}
    &\scalebox{0.78}{0.439}
    &\scalebox{0.78}{0.416} 
    &\scalebox{0.78}{0.415} 
    &\scalebox{0.78}{0.462}
    &\scalebox{0.78}{0.430} \\ 
    
    \cmidrule(lr){2-26}
    & \scalebox{0.78}{Avg} 
    & \scalebox{0.78}{0.265} 
    &\scalebox{0.78}{0.324} 
    & \secondres{\scalebox{0.78}{0.258}} 
    &\secondres{\scalebox{0.78}{0.320}} 
    & \boldres{\scalebox{0.78}{0.254}} 
    &\boldres{\scalebox{0.78}{0.315}}
    & \scalebox{0.78}{0.317} 
    & \scalebox{0.78}{0.365} 
    & \scalebox{0.78}{0.316} 
    & \scalebox{0.78}{0.361} 
    & \scalebox{0.78}{0.288}
    & \scalebox{0.78}{0.344} 
    &\scalebox{0.78}{0.273}
    & \scalebox{0.78}{0.336}
    & \scalebox{0.78}{0.311}
    &\scalebox{0.78}{0.337}
    &\scalebox{0.78}{0.329} 
    &\scalebox{0.78}{0.343} 
    & \scalebox{0.78}{0.310} 
    &\scalebox{0.78}{0.350} 
    &\scalebox{0.78}{0.295} 
    &\scalebox{0.78}{0.338} 
    &\scalebox{0.78}{0.328} 
    &\scalebox{0.78}{0.346} \\
    \midrule
    
    \multirow{5}{*}{\rotatebox{90}{\scalebox{0.95}{ETTh1}}}
    &  \scalebox{0.78}{96} 
    &\boldres{\scalebox{0.78}{0.341}} 
    &\secondres{\scalebox{0.78}{0.381}} 
    &\scalebox{0.78}{0.348} 
    &\scalebox{0.78}{0.385} 
    &\secondres{\scalebox{0.78}{0.346}} 
    &\scalebox{0.78}{0.383} 
    & \scalebox{0.78}{0.357} 
    & \scalebox{0.78}{0.381} 
    & \scalebox{0.78}{0.350} 
    & \scalebox{0.78}{0.382} 
    & \scalebox{0.78}{0.349} 
    & \boldres{{\scalebox{0.78}{0.379}}}
    & \scalebox{0.78}{0.369} 
    & \scalebox{0.78}{0.391} 
    & \scalebox{0.78}{0.376} 
    &\scalebox{0.78}{0.392} 
    & \scalebox{0.78}{0.381} 
    &\scalebox{0.78}{0.388} 
    &\scalebox{0.78}{0.440} 
    &\scalebox{0.78}{0.393} 
    &\scalebox{0.78}{0.441} 
    &\scalebox{0.78}{0.390} 
    &\scalebox{0.78}{0.414} 
    &\scalebox{0.78}{0.404}  \\ 
    
    & \scalebox{0.78}{192} 
    &\boldres{\scalebox{0.78}{0.381}}
    &\secondres{{\scalebox{0.78}{0.408}}}
    &\scalebox{0.78}{0.393} 
    &\scalebox{0.78}{0.418}
    &\scalebox{0.78}{0.386} 
    &\scalebox{0.78}{0.410} 
    & \secondres{{\scalebox{0.78}{0.384}}}
    & \boldres{\scalebox{0.78}{0.404}} 
    & \scalebox{0.78}{0.388}
    & \scalebox{0.78}{0.412}
    & \scalebox{0.78}{0.395} 
    & \scalebox{0.78}{0.413}  
    & \scalebox{0.78}{0.405} 
    & \scalebox{0.78}{0.413} 
    & \scalebox{0.78}{0.412} 
    &\scalebox{0.78}{0.413}  
    &\scalebox{0.78}{0.434}
    &\scalebox{0.78}{0.415}
    & \scalebox{0.78}{0.492} 
    &\scalebox{0.78}{0.426} 
    &\scalebox{0.78}{0.502} 
    &\scalebox{0.78}{0.524} 
    &\scalebox{0.78}{0.465} 
    &\scalebox{0.78}{0.434} \\

    & \scalebox{0.78}{336} 
    &\boldres{\scalebox{0.78}{0.405}} 
    &\secondres{\scalebox{0.78}{0.424}}
    &\scalebox{0.78}{0.422} 
    &\scalebox{0.78}{0.440}
    &\secondres{\scalebox{0.78}{0.410}}
    &\scalebox{0.78}{0.426} 
    & \scalebox{0.78}{0.411}
    & \scalebox{0.78}{0.434}
    & \scalebox{0.78}{0.411}
    & \scalebox{0.78}{0.430} 
    & \scalebox{0.78}{0.447} 
    & \scalebox{0.78}{0.453} 
    & \scalebox{0.78}{0.418}
    & \boldres{\scalebox{0.78}{0.423}} 
    & \scalebox{0.78}{0.433} 
    &\scalebox{0.78}{0.428}
    &\scalebox{0.78}{0.485}
    & \scalebox{0.78}{0.445}
    & \scalebox{0.78}{0.550}
    &\scalebox{0.78}{0.462}
    &\scalebox{0.78}{0.576} 
    &\scalebox{0.78}{0.467} 
    &\scalebox{0.78}{0.503} 
    &\scalebox{0.78}{0.456} \\ 
    
    & \scalebox{0.78}{720} 
    &\scalebox{0.78}{0.433} 
    &\scalebox{0.78}{0.458}
    &\scalebox{0.78}{0.481} 
    &\scalebox{0.78}{0.493}
    &\scalebox{0.78}{0.438} 
    &\scalebox{0.78}{0.459} 
    & \scalebox{0.78}{0.449}
    & \scalebox{0.78}{0.477}
    & \secondres{{\scalebox{0.78}{0.427}}}
    & \scalebox{0.78}{0.455}
    & \scalebox{0.78}{0.457} 
    & \scalebox{0.78}{0.462} 
    & \boldres{{\scalebox{0.78}{0.423}}} 
    & \boldres{\scalebox{0.78}{0.441}} 
    & \scalebox{0.78}{0.447} 
    &\secondres{\scalebox{0.78}{0.444}} 
    &\scalebox{0.78}{0.611} 
    &\scalebox{0.78}{0.510} 
    & \scalebox{0.78}{0.882}
    &\scalebox{0.78}{0.591}
    &\scalebox{0.78}{0.835} 
    &\scalebox{0.78}{0.583} 
    &\scalebox{0.78}{0.511}
    &\scalebox{0.78}{0.481}  \\ 
    
    \cmidrule(lr){2-26}
     &  \scalebox{0.78}{Avg}  
     & \boldres{\scalebox{0.78}{0.390}} 
     & \secondres{\scalebox{0.78}{0.418}}
     & \scalebox{0.78}{0.411} 
     & \scalebox{0.78}{0.434} 
     & \scalebox{0.78}{0.395} 
     & \scalebox{0.78}{0.420} 
     & \scalebox{0.78}{0.400} 
     & \scalebox{0.78}{0.424} 
     & \secondres{\scalebox{0.78}{0.394}} 
     & \scalebox{0.78}{0.419} 
     & \scalebox{0.78}{0.412} 
     & \scalebox{0.78}{0.426} 
     & \scalebox{0.78}{0.404} 
     & \boldres{\scalebox{0.78}{0.417}}
     & \scalebox{0.78}{0.417} 
     & \scalebox{0.78}{0.419} 
     & \scalebox{0.78}{0.480} 
     & \scalebox{0.78}{0.439} 
     & \scalebox{0.78}{0.591} 
     & \scalebox{0.78}{0.468} 
     & \scalebox{0.78}{0.588} 
     & \scalebox{0.78}{0.466} 
     & \scalebox{0.78}{0.473} 
     & \scalebox{0.78}{0.443} \\    \midrule

    \multirow{5}{*}{\rotatebox{90}{\scalebox{0.95}{ETTh2}}}
    & \scalebox{0.78}{96} 
    &\scalebox{0.78}{0.272} 
    &\secondres{\scalebox{0.78}{0.332}}
    &\secondres{\scalebox{0.78}{0.271}}
    &\scalebox{0.78}{0.333}
    &\boldres{\scalebox{0.78}{0.269}}
    &\boldres{\scalebox{0.78}{0.330}}
    & \scalebox{0.78}{0.305}
    & \scalebox{0.78}{0.359}
    & \scalebox{0.78}{0.302}
    & \scalebox{0.78}{0.354}
    & \scalebox{0.78}{0.292}
    & \scalebox{0.78}{0.352} 
    & \scalebox{0.78}{0.283}
    & \scalebox{0.78}{0.342}
    & \scalebox{0.78}{0.294}
    & \boldres{\scalebox{0.78}{0.330}} 
    &\scalebox{0.78}{0.296} 
    &\boldres{\scalebox{0.78}{0.330}} 
    &\scalebox{0.78}{0.308} 
    &\scalebox{0.78}{0.343} 
    &\scalebox{0.78}{0.320} 
    &\scalebox{0.78}{0.345} 
    &\scalebox{0.78}{0.315} 
    &\scalebox{0.78}{0.349} \\ 
    
    & \scalebox{0.78}{192} 
    &\scalebox{0.78}{0.329} 
    &\scalebox{0.78}{0.374}
    &\secondres{\scalebox{0.78}{0.327}}
    &\scalebox{0.78}{0.376}
    &\boldres{\scalebox{0.78}{0.325}}
    &\secondres{\scalebox{0.78}{0.373}}
    & \scalebox{0.78}{0.351}
    & \scalebox{0.78}{0.386}
    &\scalebox{0.78}{0.364}
    & \scalebox{0.78}{0.385}
    & \scalebox{0.78}{0.347}
    & \scalebox{0.78}{0.379} 
    & \scalebox{0.78}{0.340}
    & \scalebox{0.78}{0.379}
    & \scalebox{0.78}{0.365}
    & \scalebox{0.78}{0.375}
    &\scalebox{0.78}{0.361} 
    &\boldres{\scalebox{0.78}{0.371}} 
    & \scalebox{0.78}{0.384}
    &\scalebox{0.78}{0.392}
    &\scalebox{0.78}{0.406} 
    &\scalebox{0.78}{0.399} 
    &\scalebox{0.78}{0.388} 
    &\scalebox{0.78}{0.395} \\ 
    
    & \scalebox{0.78}{336} 
    &\secondres{\scalebox{0.78}{0.357}}
    &\secondres{\scalebox{0.78}{0.399}}
    &\boldres{\scalebox{0.78}{0.354}}
    &\scalebox{0.78}{0.402}
    &\boldres{\scalebox{0.78}{0.354}}
    &\scalebox{0.78}{0.400} 
    & \scalebox{0.78}{0.391}
    & \scalebox{0.78}{0.418}
    & \scalebox{0.78}{0.417}
    &\scalebox{0.78}{0.425}
    & \scalebox{0.78}{0.406} 
    & \scalebox{0.78}{0.419} 
    & \scalebox{0.78}{0.366}
    & \scalebox{0.78}{0.400}
    &  \scalebox{0.78}{0.376}
    & \boldres{\scalebox{0.78}{0.390}} 
    &\scalebox{0.78}{0.390} 
    &\boldres{\scalebox{0.78}{0.390}} 
    & \scalebox{0.78}{0.429}
    &\scalebox{0.78}{0.430} 
    &\scalebox{0.78}{0.492} 
    &\scalebox{0.78}{0.453} 
    &{\scalebox{0.78}{0.422}} 
    &\scalebox{0.78}{0.427}\\ 
    
    & \scalebox{0.78}{720} 
    &\scalebox{0.78}{0.401} 
    &\scalebox{0.78}{0.442} 
    &\boldres{\scalebox{0.78}{0.381}}
    &\scalebox{0.78}{0.435} 
    &\secondres{\scalebox{0.78}{0.389}}
    &\scalebox{0.78}{0.443} 
    & \scalebox{0.78}{0.419}
    & \scalebox{0.78}{0.454}
    &\scalebox{0.78}{0.537}
    & \scalebox{0.78}{0.496}
    & \scalebox{0.78}{0.439} 
    & \scalebox{0.78}{0.447} 
    & \scalebox{0.78}{0.397}
    & \secondres{\scalebox{0.78}{0.431}} 
    & \scalebox{0.78}{0.416}
    & \scalebox{0.78}{0.433}
    &\scalebox{0.78}{0.423} 
    &\boldres{\scalebox{0.78}{0.418}} 
    & \scalebox{0.78}{0.501}
    &\scalebox{0.78}{0.477}
    &\scalebox{0.78}{0.603} 
    &\scalebox{0.78}{0.511} 
    &\scalebox{0.78}{0.443} 
    &\scalebox{0.78}{0.454} \\ 
    
    \cmidrule(lr){2-26}
     &  \scalebox{0.78}{Avg}  
     & {\scalebox{0.78}{0.340}} 
     & \scalebox{0.78}{0.387} 
     & \boldres{\scalebox{0.78}{0.333}} 
     & \scalebox{0.78}{0.387} 
     & \secondres{\scalebox{0.78}{0.334}}
     & \scalebox{0.78}{0.387} 
     & \scalebox{0.78}{0.366} 
     & \scalebox{0.78}{0.404} 
     & \scalebox{0.78}{0.405} 
     & \scalebox{0.78}{0.415} 
     & \scalebox{0.78}{0.371} 
     & \scalebox{0.78}{0.399} 
     & \scalebox{0.78}{0.347} 
     & \scalebox{0.78}{0.388} 
     & \scalebox{0.78}{0.362} 
     & \secondres{\scalebox{0.78}{0.382}}
     & \scalebox{0.78}{0.367} 
     & \boldres{\scalebox{0.78}{0.377}}
     & \scalebox{0.78}{0.405} 
     & \scalebox{0.78}{0.410} 
     & \scalebox{0.78}{0.455} 
     & \scalebox{0.78}{0.427}
     & \scalebox{0.78}{0.392}
     & \scalebox{0.78}{0.406}
     \\    
     \midrule
    
    \multirow{5}{*}{\rotatebox{90}{\scalebox{0.95}{ECL}}} 
    &  \scalebox{0.78}{96} 
    &\scalebox{0.78}{0.134} 
    &\scalebox{0.78}{0.231} 
    &\secondres{\scalebox{0.78}{0.132}}
    &\secondres{\scalebox{0.78}{0.229}}
    &\boldres{\scalebox{0.78}{0.130}}
    &\boldres{\scalebox{0.78}{0.227}}
    & \scalebox{0.78}{-}
    & \scalebox{0.78}{-} 
    & \scalebox{0.78}{-} 
    & \scalebox{0.78}{-}
    & \scalebox{0.78}{-}
    & \scalebox{0.78}{-}
    & \scalebox{0.78}{0.141}
    & \scalebox{0.78}{0.237}
    & \scalebox{0.78}{0.160}
    &\scalebox{0.78}{0.250} 
    &\scalebox{0.78}{0.153} 
    &\scalebox{0.78}{0.241} 
    &\scalebox{0.78}{0.154} 
    &\scalebox{0.78}{0.231}
    &\scalebox{0.78}{0.152}
    &\secondres{\scalebox{0.78}{0.229}} 
    &\scalebox{0.78}{-}
    &\scalebox{0.78}{-} \\

    & \scalebox{0.78}{192} 
    &\scalebox{0.78}{0.154} 
    &\scalebox{0.78}{0.251}
    &\secondres{\scalebox{0.78}{0.152}}
    &\secondres{\scalebox{0.78}{0.250}}
    &\boldres{\scalebox{0.78}{0.150}}
    &\boldres{\scalebox{0.78}{0.247}}
    & \scalebox{0.78}{-}
    & \scalebox{0.78}{-}
    & \scalebox{0.78}{-} 
    & \scalebox{0.78}{-}
    & \scalebox{0.78}{-}
    & \scalebox{0.78}{-}
    & \scalebox{0.78}{0.159}
    & \scalebox{0.78}{0.254}
    & \scalebox{0.78}{0.175}
    &\scalebox{0.78}{0.263} 
    &\scalebox{0.78}{0.169}
    &\scalebox{0.78}{0.255}
    & \scalebox{0.78}{0.179} 
    &\scalebox{0.78}{0.254}
    &\scalebox{0.78}{0.172}
    &\secondres{\scalebox{0.78}{0.250}}
    &\scalebox{0.78}{-} 
    &\scalebox{0.78}{-} \\ 
    
    & \scalebox{0.78}{336} 
    &\scalebox{0.78}{0.174} 
    &\scalebox{0.78}{0.271}
    &\secondres{\scalebox{0.78}{0.173}}
    &\secondres{\scalebox{0.78}{0.271}}
    &\boldres{\scalebox{0.78}{0.170}}
    &\boldres{\scalebox{0.78}{0.268}}
    & \scalebox{0.78}{-} 
    & \scalebox{0.78}{-} 
    & \scalebox{0.78}{-}
    & \scalebox{0.78}{-}
    & \scalebox{0.78}{-}
    & \scalebox{0.78}{-}
    & \scalebox{0.78}{0.177}
    & \scalebox{0.78}{0.272}
    & \scalebox{0.78}{0.187}
    &\scalebox{0.78}{0.277}
    &\scalebox{0.78}{0.187} 
    &\scalebox{0.78}{0.273}
    & \scalebox{0.78}{0.214} 
    &\scalebox{0.78}{0.284} 
    &\scalebox{0.78}{0.203}
    &\scalebox{0.78}{0.276} 
    &\scalebox{0.78}{-} 
    &\scalebox{0.78}{-}  \\ 
    
    & \scalebox{0.78}{720}
    &\secondres{\scalebox{0.78}{0.215}}
    &\boldres{\scalebox{0.78}{0.307}}
    &\scalebox{0.78}{0.218} 
    &\scalebox{0.78}{0.311}
    &\boldres{\scalebox{0.78}{0.214}}
    &\boldres{\scalebox{0.78}{0.307}}
    & \scalebox{0.78}{-} 
    & \scalebox{0.78}{-} 
    & \scalebox{0.78}{-}
    & \scalebox{0.78}{-}
    & \scalebox{0.78}{-}
    & \scalebox{0.78}{-}
    & \scalebox{0.78}{0.219}
    & \secondres{\scalebox{0.78}{0.308}} 
    &\scalebox{0.78}{0.228}
    &\scalebox{0.78}{0.309}
    &\scalebox{0.78}{0.237} 
    &\scalebox{0.78}{0.313} 
    & \scalebox{0.78}{0.311} 
    &\scalebox{0.78}{0.346} 
    &\scalebox{0.78}{0.289}
    &\scalebox{0.78}{0.337}
    &\scalebox{0.78}{-} 
    &\scalebox{0.78}{-} \\ 
    
    \cmidrule(lr){2-26}
     &  \scalebox{0.78}{Avg}  
     & \secondres{\scalebox{0.78}{0.169}} 
     & \secondres{\scalebox{0.78}{0.265}}
     &\secondres{\scalebox{0.78}{0.169}} 
     & \secondres{\scalebox{0.78}{0.265}}
     & \boldres{\scalebox{0.78}{0.166}} 
     & \boldres{\scalebox{0.78}{0.262}}
     & \scalebox{0.78}{-} 
     & \scalebox{0.78}{-} 
     & \scalebox{0.78}{-} 
     & \scalebox{0.78}{-} 
     & \scalebox{0.78}{-} 
     & \scalebox{0.78}{-} 
     & \scalebox{0.78}{0.174} 
     & \scalebox{0.78}{0.278} 
     & \scalebox{0.78}{0.187} 
     & \scalebox{0.78}{0.274} 
     & \scalebox{0.78}{0.186} 
     & \scalebox{0.78}{0.270} 
     & \scalebox{0.78}{0.214} 
     & \scalebox{0.78}{0.278} 
     & \scalebox{0.78}{0.204} 
     & \scalebox{0.78}{0.273} 
     & \scalebox{0.78}{-} 
     & \scalebox{0.78}{-} \\

    \midrule
    \multirow{5}{*}{\rotatebox{90}{\scalebox{0.95}{Weather}}} 
    & \scalebox{0.78}{96} 
    &\secondres{\scalebox{0.78}{0.158}}
    &\secondres{\scalebox{0.78}{0.206}}
    &\boldres{\scalebox{0.78}{0.157}}
    &\boldres{\scalebox{0.78}{0.205}}
    &\boldres{\scalebox{0.78}{0.157}}
    &\scalebox{0.78}{0.208} 
    & \scalebox{0.78}{0.160} 
    & \scalebox{0.78}{0.214} 
    & {\scalebox{0.78}{0.159}} 
    & {\scalebox{0.78}{0.213}} 
    & \boldres{\scalebox{0.78}{0.157}}
    & {\scalebox{0.78}{0.211}}  
    & \scalebox{0.78}{0.171} 
    & {\scalebox{0.78}{0.225}} 
    & {\scalebox{0.78}{0.220}} 
    &{\scalebox{0.78}{0.217}} 
    & \scalebox{0.78}{0.199} 
    &{\scalebox{0.78}{0.211}} 
    & \scalebox{0.78}{0.203} 
    &\scalebox{0.78}{0.238} 
    & {\scalebox{0.78}{0.194}} 
    &{\scalebox{0.78}{0.235}} 
    & \scalebox{0.78}{-} 
    &\scalebox{0.78}{-} \\ 
    
    & \scalebox{0.78}{192} 
    &\boldres{\scalebox{0.78}{0.205}}
    &\secondres{\scalebox{0.78}{0.253}}
    &\boldres{\scalebox{0.78}{0.205}}
    &\boldres{\scalebox{0.78}{0.251}}
    &\secondres{\scalebox{0.78}{0.207}}
    &\scalebox{0.78}{0.256} 
    & {\scalebox{0.78}{0.210}}
    & \scalebox{0.78}{0.260} 
    & \scalebox{0.78}{0.215} 
    & \scalebox{0.78}{0.266}
    & {\scalebox{0.78}{0.208}} 
    & {\scalebox{0.78}{0.256}} 
    & \scalebox{0.78}{0.221} 
    & {\scalebox{0.78}{0.271}} 
    & {\scalebox{0.78}{0.271}}
    &{\scalebox{0.78}{0.259}}  
    & \scalebox{0.78}{0.246} 
    &\boldres{\scalebox{0.78}{0.251}} 
    &\scalebox{0.78}{0.256} 
    &\scalebox{0.78}{0.290} 
    & \scalebox{0.78}{0.249} 
    &\scalebox{0.78}{0.285} 
    & \scalebox{0.78}{-} 
    &\scalebox{0.78}{-} \\ 
    
    & \scalebox{0.78}{336} 
    &\secondres{\scalebox{0.78}{0.254}}
    &\secondres{\scalebox{0.78}{0.290}}
    &\boldres{\scalebox{0.78}{0.253}}
    &\boldres{\scalebox{0.78}{0.289}}
    &\scalebox{0.78}{0.259} 
    &\scalebox{0.78}{0.295} 
    & \scalebox{0.78}{0.274} 
    & \scalebox{0.78}{0.309}
    & \scalebox{0.78}{0.291} 
    & {\scalebox{0.78}{0.322}} 
    & {\scalebox{0.78}{0.255}} 
    & \secondres{\scalebox{0.78}{0.290}}
    & {\scalebox{0.78}{0.274}} 
    & {\scalebox{0.78}{0.311}} 
    & {\scalebox{0.78}{0.286}} 
    &{\scalebox{0.78}{0.297}} 
    & {\scalebox{0.78}{0.274}} 
    &{\scalebox{0.78}{0.291}} 
    & \scalebox{0.78}{0.314} 
    &\scalebox{0.78}{0.336}
    & \scalebox{0.78}{0.302} 
    &\scalebox{0.78}{0.327}
    & \scalebox{0.78}{-} 
    &\scalebox{0.78}{-}\\ 
    
    & \scalebox{0.78}{720} 
    &\boldres{\scalebox{0.78}{0.315}}
    &\boldres{\scalebox{0.78}{0.336}}
    &\secondres{\scalebox{0.78}{0.320}}
    &\boldres{\scalebox{0.78}{0.336}}
    &\scalebox{0.78}{0.327} 
    &\secondres{\scalebox{0.78}{0.342}}
    & \scalebox{0.78}{0.418} 
    & \scalebox{0.78}{0.405} 
    & \scalebox{0.78}{0.415} 
    & {\scalebox{0.78}{0.400}} 
    & \scalebox{0.78}{0.405} 
    & \scalebox{0.78}{0.397} 
    & \scalebox{0.78}{0.356} 
    & {\scalebox{0.78}{0.370}} 
    & \scalebox{0.78}{0.373} 
    &{\scalebox{0.78}{0.354}} 
    & {\scalebox{0.78}{0.337}} 
    &{\scalebox{0.78}{0.340}} 
    & {\scalebox{0.78}{0.397}} 
    &\scalebox{0.78}{0.396} 
    & \scalebox{0.78}{0.372} 
    &\scalebox{0.78}{0.378} 
    & \scalebox{0.78}{-} 
    &\scalebox{0.78}{-} \\

    \cmidrule(lr){2-26}
     &  \scalebox{0.78}{Avg}  
     & \boldres{\scalebox{0.78}{0.233}} 
     & \secondres{\scalebox{0.78}{0.271}}
     & \secondres{\scalebox{0.78}{0.234}}
     & \boldres{\scalebox{0.78}{0.270}}
     & \scalebox{0.78}{0.238} 
     & \scalebox{0.78}{0.275}
     & \scalebox{0.78}{0.265}
     & \scalebox{0.78}{0.297} 
     & \scalebox{0.78}{0.270} 
     & \scalebox{0.78}{0.300} 
     & \scalebox{0.78}{0.256}
     & \scalebox{0.78}{0.288}
     & \scalebox{0.78}{0.256} 
     & \scalebox{0.78}{0.294}
     & \scalebox{0.78}{0.287}
     & \scalebox{0.78}{0.281} 
     & \scalebox{0.78}{0.264}
     & \scalebox{0.78}{0.273}
     & \scalebox{0.78}{0.292} 
     & \scalebox{0.78}{0.315} 
     & \scalebox{0.78}{0.279} 
     & \scalebox{0.78}{0.306} 
     & \scalebox{0.78}{-} 
     & \scalebox{0.78}{-} \\
    
    \midrule
    \multicolumn{2}{c|}{\textbf{\scalebox{0.78}{{$1^{\text{st}}$ Count}}}} &\scalebox{0.78}{7} 
     &\scalebox{0.78}{2} 
     & \secondres{\scalebox{0.78}{{8}}}
     & \scalebox{0.78}{{5}} 
     & \boldres{\scalebox{0.78}{16}}
     & \boldres{\scalebox{0.78}{{16}}}
     &\scalebox{0.78}{{0}} 
     &\scalebox{0.78}{{1}} 
     & \scalebox{0.78}{0} 
     & \scalebox{0.78}{0} 
     & \scalebox{0.78}{2} 
     & \scalebox{0.78}{1} 
     & \scalebox{0.78}{{1}} 
     & \scalebox{0.78}{3}
     & \scalebox{0.78}{0} 
     & \scalebox{0.78}{{2}} 
     & \scalebox{0.78}{0} 
     & \secondres{\scalebox{0.78}{6}}
     & \scalebox{0.78}{0} 
     & \scalebox{0.78}{0} 
     & \scalebox{0.78}{0} 
     & \scalebox{0.78}{0} 
     & \scalebox{0.78}{0} 
     & \scalebox{0.78}{0} \\ 
    \bottomrule
  \end{tabular}
    \end{small}
        \begin{tablenotes}
        \footnotesize
        \item[] $\ast$ Traffic~\cite{trafficdata} is not evaluated because it is included in the pre-training datasets of these time series foundation models.
  \end{tablenotes}
  \end{threeparttable}
\vspace{-5pt}
\end{table}

\subsection{Showcases of Generative Forecasters and Deterministic Forecasters}\label{app:showcase_compare}
\begin{figure*}[ht]
\begin{center}
    \center{\includegraphics[width=\textwidth]{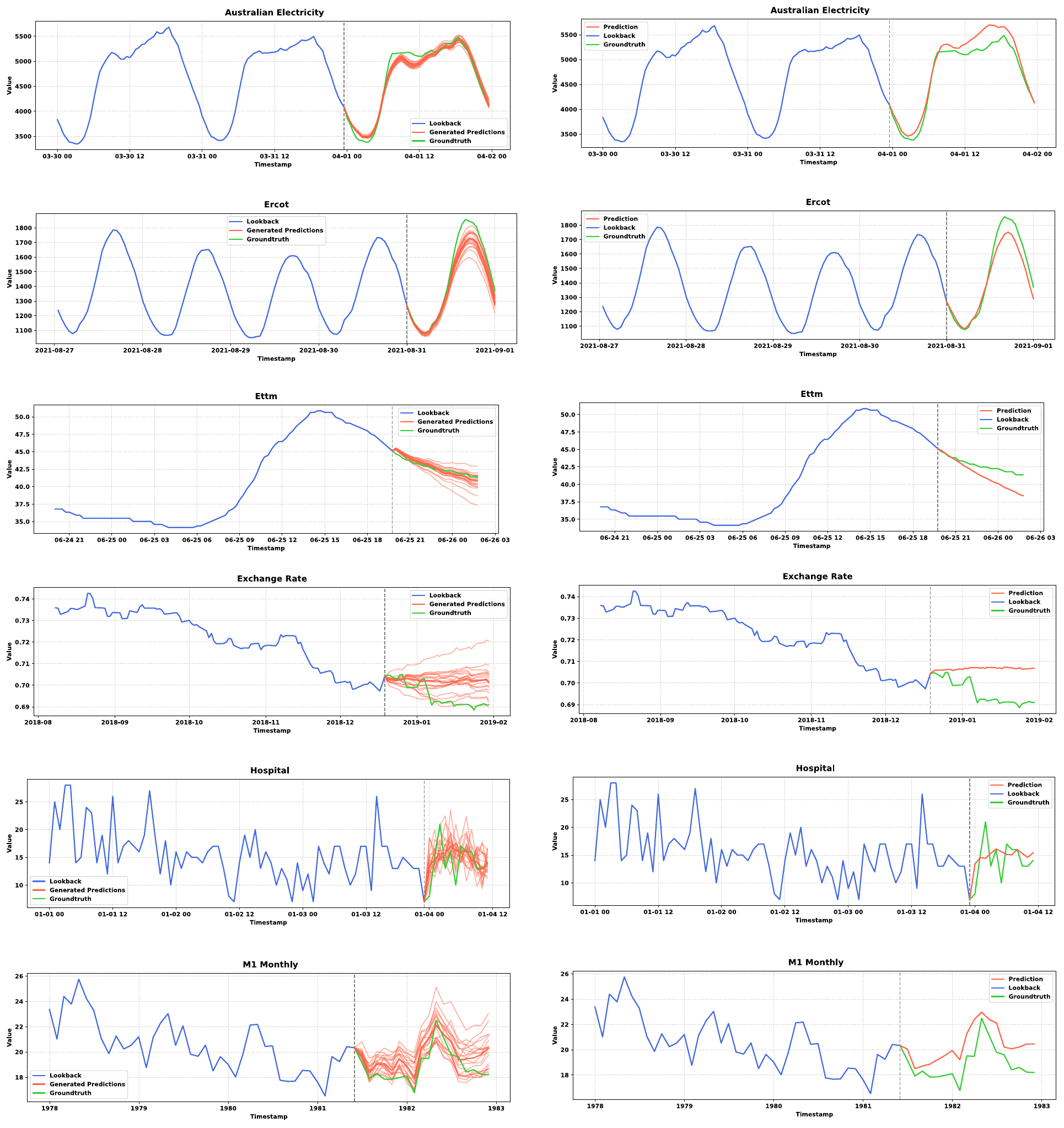}}
    \caption{Showcases of Sundial (Left) and the same Transformer backbone pre-trained by MSE Loss (Right). MSE Loss optimizes a deterministic forecaster: given a lookback series, the model can only produce one prediction as the estimation of mean values. This objective may fail to accommodate divergent future variations during large-scale pre-training, leading to mode collapse and over-smooth results (as illustrated in the fourth row). TimeFlow optimizes a generative forecaster: it generates various possibilities observed in the pre-training dataset. Based on these raw predictions, we can estimate the underlying complicated distribution and different statistics. Besides, the greater concentration in generated predictions, the higher the model's confidence in its predictions.}
	\label{fig:showcases_compare1}
\end{center}
\end{figure*}

\begin{figure*}[ht]
\begin{center}
    \center{\includegraphics[width=\textwidth]{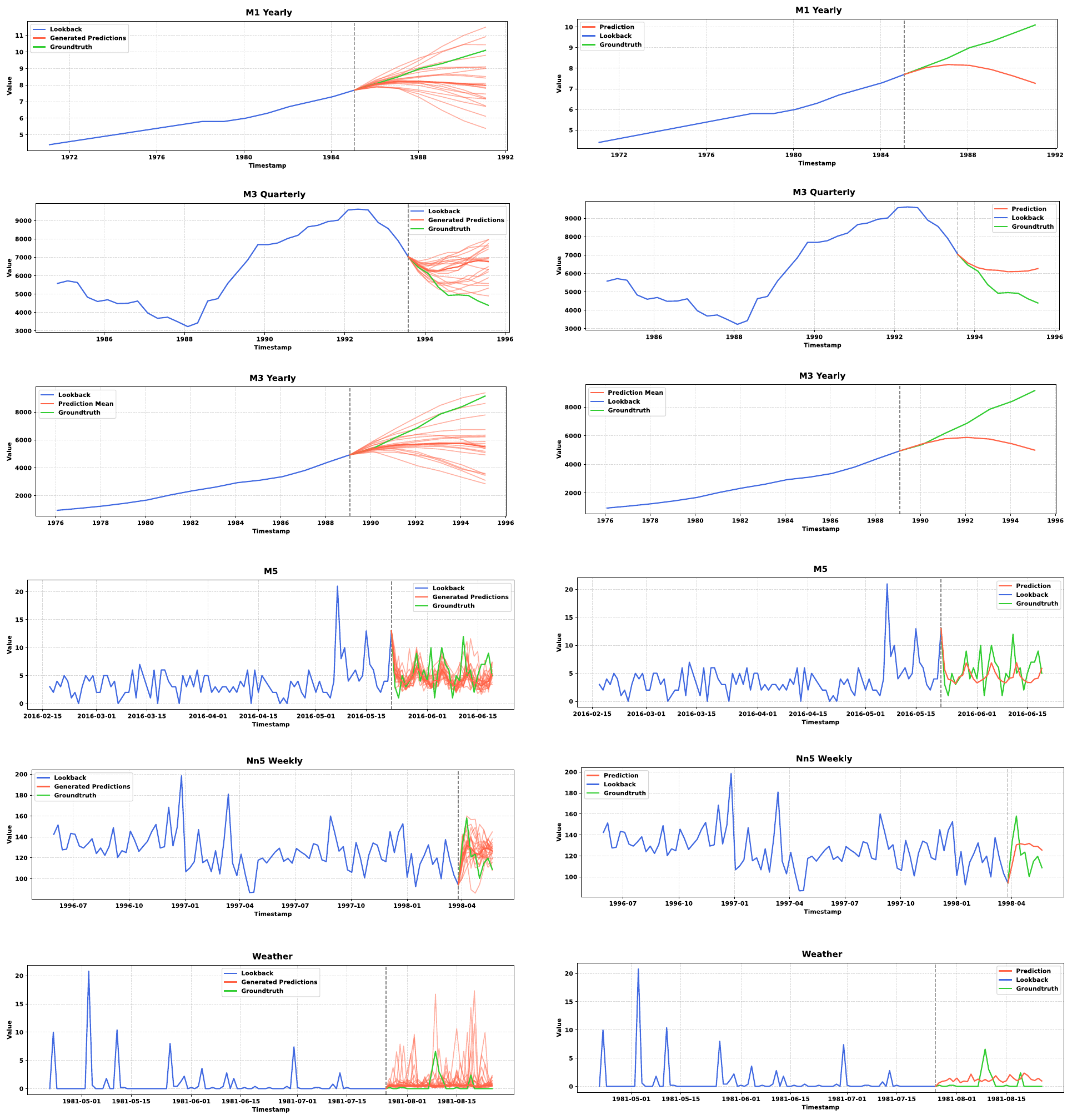}}
    \caption{Supplementary showcases of Sundial (Left) and the same Transformer backbone pre-trained by MSE Loss (Right).}
	\label{fig:showcases_compare2}
\end{center}
\end{figure*}

As we introduce generative modeling in time series foundation models, we compare zero-shot forecasting showcases from two types of models, including (1) Sundial, a generative forecaster pre-trained by TimeFlow, which can predict multiple future possibilities based on a lookback series. (2) Using the same backbone and TimeBench, a Transformer pre-trained by MSE Loss. As a deterministic forecaster, the model can only output the mean prediction. As depicted in Figure~\ref{fig:showcases_compare1}-\ref{fig:showcases_compare2}, the unimodal Gaussian prior specified by MSE can be infeasible to handle large-scale pre-training, manifested as sometimes over-smooth predictions in downstream forecasting tasks. Therefore, we hope this work can inspire future paradigms for pre-training time series foundation models and enhance their applicability to real-world scenarios.

\section{Limitations}
Our models represent an initial effort to incorporate generative modeling into time series foundation models, which enables pre-training on heterogeneous time series without specifying any prior distribution. This approach mitigates mode collapse in representation learning and generates a diverse range of probable predictions compared to previous deterministic forecasters. Despite significant progress in enlarging model capacity, the Sundial family may still face hallucinations. The performance on very high-frequency data is not guaranteed, since TimeBench contains many middle- and low-frequency time series. Therefore, an important future direction is to generalize Sundial to multi-scale time series. This situation may also indicate new opportunities during inference. As we only adopt a na\"ive sampling strategy that begins with random Gaussian noise, it leaves much room for future improvement in sampling strategy and post-processing, such as frequency normalization.

Another aspect of future development lies in model adaptation. Sundial is pre-trained in a univariate approach to address the discrepancy in variate numbers, which prevents it from explicitly utilizing variate correlations or covariate information. As an increasing number of studies address 2D dimensionality, multivariate pre-training is likely to be conducted for domain-specific time series foundation models. Lastly, while autoregressive models provide flexibility in the input context length, multiple steps of autoregression may still lead to over-smooth predictions and unreliable results. 

\section{Societal Impacts}

\subsection{Real-World Applications} 
In this work, we present Sundial, a family of time series foundation models  to facilitate out-of-the-box forecasting. Our models employ native tokenization for continuous-valued time series and incorporate a flexible training objective, proposed as TimeFlow Loss, to enable probabilistic forecasting. With an unprecedented model capacity and a trillion-scale dataset, our models can be used directly or adapted for various forecasting scenarios, such as energy planning, weather forecasting, and financial risk prevention. With multiple predictions generation and a just-in-time inference speed, our model enhances the reliability of decision-making and streamlines the forecasting pipeline for practitioners. This paper primarily focuses on scientific research and does not present any evident negative social impact.

\subsection{Academic Research} 
We curate TimeBench, a trillion-level time series dataset for pre-training foundation models for time series analysis, which we believe will be beneficial to the research community. Technically, we propose a TimeFlow Loss to facilitate the learning of flexible next-patch distributions. Conditioned on the lookback representations acquired by autoregressive Transformers, our model is endowed with a novel generative capability for probabilistic forecasting, enhancing representation learning of Transformers without the need for discrete tokenization. Through pre-training on an unprecedented scale, we identify subtle scalability bottlenecks that are not solely attributable to architectural design but are predominantly influenced by the training objectives of foundation models. The proposed TimeFlow Loss applied to autoregressive and generative models may provide insights for the future development of time series foundation models.

\end{document}